\documentclass[review]{elsarticle}

\usepackage{amsfonts}
\usepackage{amsmath}
\usepackage{amssymb}
\usepackage{xcolor}

\DeclareMathOperator{\argmin}{argmin}

\makeatletter
\def\bs{\expandafter\@gobble\string\\}
\def\lb{\expandafter\@gobble\string\{}
\def\rb{\expandafter\@gobble\string\}}
\def\@pdfauthor{A.Laflaqui\`ere}
\def\@pdftitle{Discovering space - Grounding spatial topology and metric regularity in a naive agent's sensorimotor experience}
\def\@pdfsubject{Discovering space - Grounding spatial topology and metric regularity in a naive agent's sensorimotor experience}
\def\@pdfkeywords{space perception, grounding, sensorimotor contingencies, developmental robotics}

\begin{document}

\title{Discovering space - Grounding spatial topology and metric regularity in a naive agent's sensorimotor experience}

\author[sbre]{Alban Laflaqui\`ere\corref{cor1}}
\ead{alaflaquiere@softbankrobotics.com}

\author[lpp]{J.Kevin O'Regan\fnref{ackngmt}}
\ead{jkevin.oregan@gmail.com}

\author[isir]{Bruno Gas}
\ead{bruno.gas@upmc.fr}

\author[lpp]{Alexander Terekhov\fnref{ackngmt}}
\ead{avterekhov@gmail.com}

\cortext[cor1]{Corresponding author}
\fntext[ackngmt]{Acknowledge support from ERC advanced grant No 323674 "FEEL"}

\address[sbre]{AI Lab, SoftBank Robotics EU, FR}
\address[lpp]{Univ. Paris 05 Descartes and LPP (CNRS UMR 8158), FR}
\address[isir]{Univ. Paris 06 UPMC and ISIR (CNRS UMR 7222), FR}


\begin{abstract}
In line with the sensorimotor contingency theory, we investigate the problem of the perception of space from a fundamental sensorimotor perspective.
Despite its pervasive nature in our perception of the world, the origin of the concept of space remains largely mysterious. For example in the context of artificial perception, this issue is usually circumvented by having engineers pre-define the spatial structure of the problem the agent has to face.
We here show that the structure of space can be autonomously discovered by a naive agent in the form of sensorimotor regularities, that correspond to so called compensable sensory experiences: these are experiences that can be generated either by the agent or its environment. By detecting such compensable experiences the agent can infer the topological and metric structure of the external space in which its body is moving.
We propose a theoretical description of the nature of these regularities and illustrate the approach on a simulated robotic arm equipped with an eye-like sensor, and which interacts with an object. Finally we show how these regularities can be used to build an internal representation of the sensor's external spatial configuration.
\end{abstract}


\maketitle


\section{Introduction}
\label{sec:Introduction}

In the last decade progress has been made in giving more autonomy to robots. These achievements are due in large part to the development of machine learning which progressively replaces formerly hand-crafted features with features autonomously extracted from data \citep{Goodfellow-et-al-2016}. 
Yet the ultimate goal of having a robot perceive its environment in a completely unsupervised way still seems far out of reach.
Current best performing technologies require human beings to explore the data, interpret it, and provide labels to the robot indicating which aspects of its perceptual content are of interest \citep{LeCun2015}. Other more exploratory approaches try to reduce the human input to a minimum by only providing the rewards associated to particular tasks \citep{levine2016end, Lillicrap2015, Mnih2015}. The resulting systems, although often impressive on specific tasks, tend to behave like black-boxes from which it is difficult to extract what has been perceived by the robot \citep{greydanus2017visualizing,Zahavy2016}, in turn making any improvement complicated. They also show fairly poor generalization performance, as knowledge acquired while solving a given task does not transfer well to other tasks \citep{pan2010survey,rusu2016progressive}.

In order to reach a perceptual autonomy on a par with the one observed in animals, we believe roboticists need to take a step back and question the nature of the artificial perception they implement in their systems. Are they building genuine perception?
To be truly autonomous, artificial perception needs to be thought of as a mechanism which empowers the agent itself instead of satisfying a performance criterion defined by an engineer.
This mechanism has to be consistent from the intrinsic perspective of the agent, grounded in its sensorimotor experience, the only source of information it has access to \citep{Hoffman2015,Hohwy2016}. It also needs to account for the different properties of perceptual experiences and explain how they are relevant for the agent.
Only then will we be able to genuinely understand how abstract perceptual concepts can be grounded in a robot's experience.

\begin{figure}[t]
\centering
\includegraphics[width=1\linewidth]{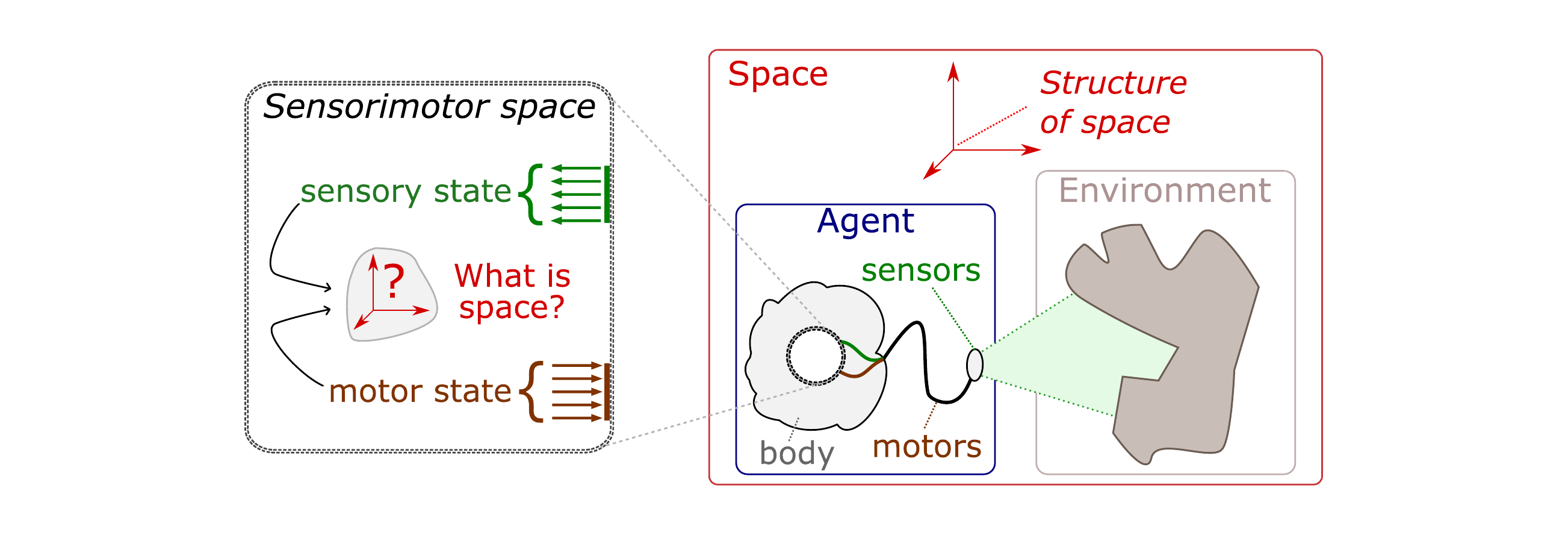}
\caption{An agent's physical body and its environment are both immersed in space. However the agent's naive "brain" (information processing system) does not have access to the properties of space, like its topology or metric, but only to an uninterpreted sensorimotor flow from which they need to be inferred.}
\label{fig:approach illustration}
\end{figure}

%

The \emph{sensorimotor contingency theory} (SMCT) \citep{ORegan2001} has furthermore investigated this question in the more general framework of a theory of perception. It notably claims that perception relies on an agent's ability to \emph{"master sensorimotor contingencies"}. The so called sensorimotor contingencies correspond to regularities induced by the world on the way the agent's actions transform its sensory inputs \citep{Laflaquiere2015}.
Beyond the philosophical discussion related to this claim, which does not belong in this paper, the SMCT standpoint gives rise to pragmatic consequences for robots. It suggests that they should be able to autonomously acquire perceptual abilities by exploring their sensorimotor space and by discovering the regularities that the world induces in their experience. In turn, perception is achieved by identifying the regularities which underlie an ongoing interaction with the world. Moreover properties of the agent's perceptual experiences would be directly connected to the properties of their underlying sensorimotor regularities \citep{ORegan2012}.
The groundbreaking viewpoint proposed by the SMCT requires a complete overhaul of theories of perception as they are currently approached in robotics and biology. To date, some perceptual concepts such as color \citep{Philipona2006,Witzel2015}, environment \citep{Hemion2017}, object \citep{laflaquiere2015objects}, visual field \citep{Laflaquiere2016}, and space \citep{Laflaquiere2012, Laflaquiere2015a, Philipona2003, Terekhov2013} have been re-addressed through the prism of sensorimotor contingencies.

The concept of \emph{space} proves to be of particular interest. Indeed space constitutes an essential component of our perception of the world. A vast majority of what we perceive, including our body, appears to us as immersed in space (see Fig.~\ref{fig:approach illustration}). Yet its fundamental nature remains mysterious \citep{kant1998critique, nicod1965geometrie, Poincare1895}.
In robotics the notion of space is traditionally considered as given and the problem of spatial knowledge acquisition bypassed by engineers. This is obviously true in industrial settings where the position and orientation of each part of the robot is analytically defined (or measured) \citep{siciliano2016springer}. It is also true in more autonomous settings, like for instance SLAM, where the spatial nature of the mapping problem and of the robot's actions are predefined \citep{Cadena2016}. More surprisingly it is even the case in most developmental experiments involving spatial knowledge. For example models for body structure discovery and forward model learning consider the spatial configuration of some part of the robot as known \citep{Dearden2005, Loviken2017, Rolf2010}.
Nonetheless completely unsupervised approaches of spatial tasks do exist \citep{Jonschkowski2015,Mirowski2016}. Unfortunately they boil down the robot's experience to an all-encompassing performance measure which rules out  the specificity of spatial experience. Even a posteriori analyses have so far been unable to unravel the intricate internal states of those agents \citep{Shwartz-Ziv1999,Zahavy2016}.
{\color{black} Finally, other approaches have previously been proposed to ground perceptive experience in an agent's sensorimotor experience \cite{choe2006motion, modayil2008initial, pierce1997map}. However they mainly focused on inferring properties of the environment without explicitly defining the specific structure of spatial experience.}

In line with the SMCT, we claim that the subjective properties we attribute to space should be reflected in the properties of the regularities it induces in our sensorimotor experiences. Space, being ubiquitous in our perception of the world, should thus significantly shape those experiences.
From our perceptual experience, we can say that space is \textit{content-independent}, which means that it does not depend on the nature of the objects it contains\footnote{Einstein would of course beg to differ as the theory of relativity describes how mass and energy distort space-time. However we are only interested here in the scale of interaction that humans and robots have with their environment, a context in which such distortion does not manifest directly.}. Space is also \textit{shared} by an agent and its environment, which means that we perceive ourselves as immersed in the same space as objects that surround us. Finally space has an isotropic structure, which means that we experience space the same way in different contexts (different environments, or different positions/orientations of the agent).

In this work our goal is to understand how these fundamental properties are grounded in the sensorimotor experience of a robot.
{\color{black} We can arguably present the field of machine learning applied to Robotics as the junction of two endeavors: understanding the data structure that needs to be captured to support intelligent behavior, and developing algorithmic solutions to capture this structure. Whereas most effort in the community is oriented towards the latter, we believe that studying the data structure that can support artificial intelligence is fundamental, and it is this approach we follow in this work. Our objective is to shed new light on the possible emergence of space perception, which might in turn inform the development of future algorithmic solutions to this problem.}
Consequently our focus is not on optimizing a controller to solve a spatial task, or on new machine learning developments, but on the fundamental definition of the sensorimotor structure underlying spatial knowledge.
This definition is directly inspired by original ideas from H.Poincar\'e who wondered more than a century ago why Euclidean geometry was natural to us \citep{Poincare1895}.
His conclusion was that space and geometry are revealed to us through the experience of sensory variations that can be generated either by a motor command of by a change in the environment.
His insights have since been developed to propose ways for a naive agent to discover the dimension of space \citep{Laflaquiere2012,Philipona2004}. Later they also initiated the sensorimotor characterization of the concepts of \emph{rigid displacement} \citep{Terekhov2014} and \emph{points of view} \citep{Laflaquiere2015a} that space enables. The results presented in this paper build on these two related concepts to propose a richer characterization of spatial experience.
{\color{black} More precisely, we show how space induces invariants in a naive agent's sensorimotor experience. We define the space-related invariants that the agent can discover, which correspond to regularities in the way sensory inputs change when the agent changes its motor configuration or when its environment changes its own configuration. Intuitively, this approach leads us to describe space as a redundancy between the agent's configuration and the environment's configuration. Furthermore, the agent can actively experience this redundancy by compensating for environmental changes through its own actions.
We show how knowledge of these space-related invariants can guide the building of an internal representation of the agent's external spatial configuration. Depending on the nature of the invariants discovered by the agent, this representation can capture both the topology and the metric regularity of the external space. We also discuss how the nature of these invariants relate to our experience of space as content-independent and shared with the environment.}

In the following sections we propose a formalism to analyze the sensorimotor experience of a naive agent as well as how space shapes this experience.
Two illustrative simulations are then presented in which a simple agent extracts space-induced sensorimotor regularities. Both are based on the detection of sensorimotor invariants, and allow the agent to build internal representations of its spatial configuration which respectively capture the topology of space and the regularity of the metric of space.
Finally the experimental and theoretical results, limitations, and potential improvements of the approach are thoroughly discussed in the last section of this paper.


\section{Problem statement}
\label{sec:Problem statement}

In this section the question of the perception of space is first addressed from a theoretical perspective by combining insights from H.Poincar\'e and the SMCT. A mathematical formalization of the problem is then proposed in order to study how space manifests itself in an agent's sensorimotor flow. Finally a simple simulated system on which the simulations of the next two sections will be evaluated is introduced.

\subsection{The sensorimotor imprint of space}
\label{sec:The sensorimotor imprint of space}

The question of the nature of spatial knowledge arises most strikingly when taking the standpoint of a naive agent with no a priori knowledge about its body, its sensors, nor the external world (space included). The sensory flow produced by typical sensors like cameras and microphones does not contain direct spatial information such as the agent's position and orientation. Besides, even assuming that the sensory flow contains such specific information, thanks to the use of dedicated sensors like gyroscopes or GPS, the agent still needs to discover what distinguishes these sensory inputs from the non-spatial information provided by other inputs. One thus has to specify in what way spatial information differs from non-spatial information, and why such a distinction is valuable and immediately useful to the agent.

The experience of a naive agent consists in the uninterpreted incoming sensory flow from its sensors and the uninterpreted motor flow it can send to its motors. This sensorimotor experience constitutes the "inner world" the agent lives in (see Fig.~\ref{fig:approach illustration}). It is obviously different from the external world the agent's physical body is embedded in and where space seems to exist as an omnipresent persistent frame.
At any time, the agent is in a certain sensorimotor state, a point in its internal sensorimotor space, that an external observer would associate with a rich description of the agent's configuration and of the content of its environment.
It is important to notice that a sensorimotor state, or its description by an observer, is not "spatial" in itself. More precisely, it is the result of a multitude of parameters, some spatial (like the positions of objects around the agent) and some non-spatial (like their colors). However \emph{changes} in sensorimotor states can be purely spatial, like for instance an object changing position without changing its color. Sensorimotor changes are thus the means by which the agent can isolate space from other properties of the world.

\begin{figure}[t]
\centering
\includegraphics[width=1\linewidth]{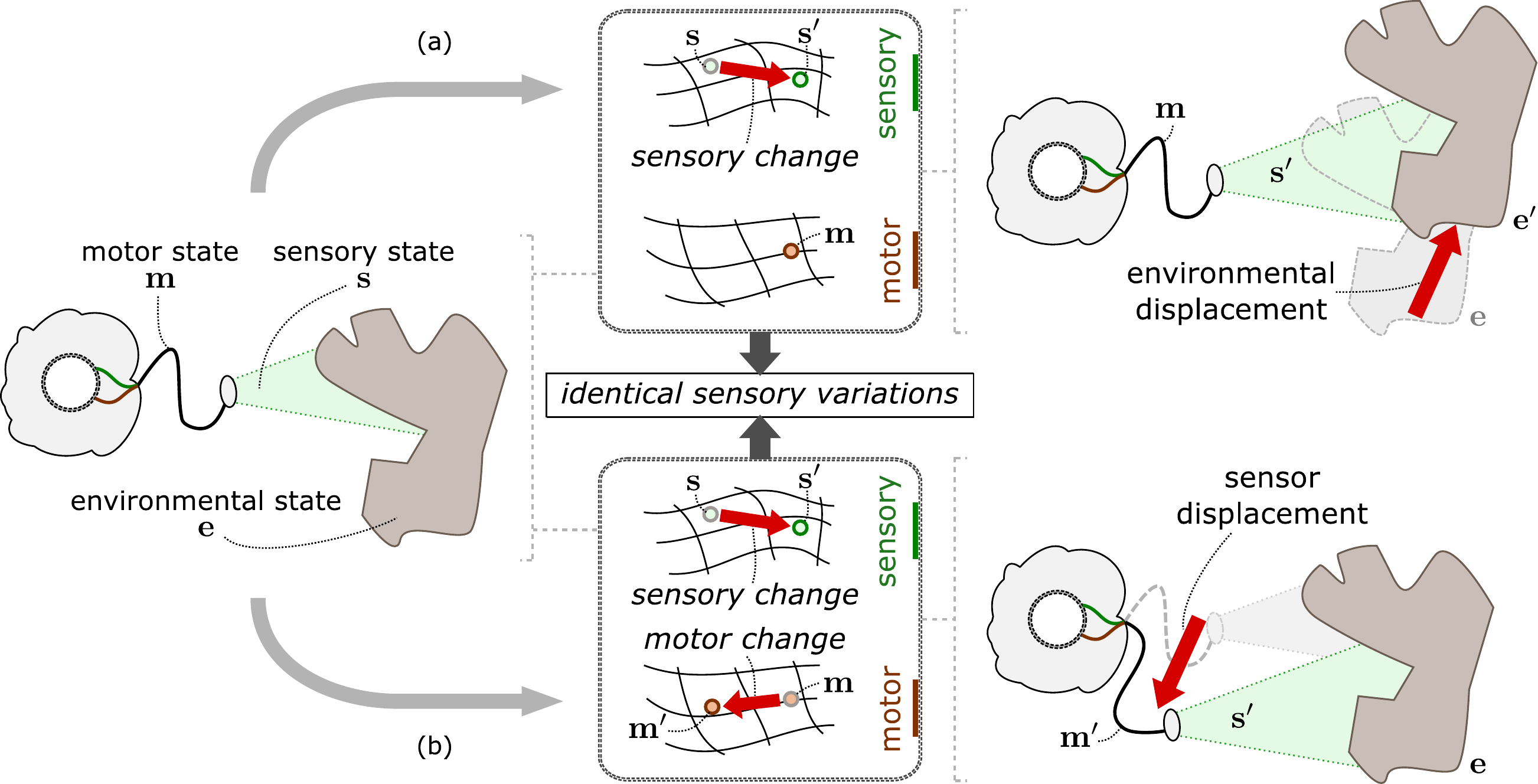} 
\caption{{\color{black}Illustration of an agent-environment system.} Displacements in the system have the particular property of being redundant from a sensory perspective: the sensory change generated by a displacement of the environment (a) can equivalently be produced by a displacement of the agent's sensor (b), and vice versa.}
\label{fig:redundancy}
\end{figure}

The sensorimotor changes an agent can undergo are of different kinds. For an external observer, either the agent or its environment can be the source of the change, and in each case the change can be either spatial or non-spatial. We later refer to non-spatial changes as \emph{state changes}. For instance, a spatial change of the agent could be the displacement of one of its body parts, a state change due to the agent could be the modification of a sensor sensitivity, a spatial change of the environment could be the displacement of an object, and an environmental state change could be a change of the object's color or electrical conductivity.
For the naive agent with no prior knowledge about space, there is only one distinction: some sensory changes occur when motor commands are emitted (active kind), while others occur while the motor state is static (passive kind). The agent thus needs to rely on an additional property to distinguish between spatial and state changes in both active and passive cases.

Intuitions formulated by H.Poincar\'e as early as the late 19th century suggest that such a property does exist in the form of sensory redundancy between the agent and its environment \citep{Poincare1895}.
Indeed while exploring its inner sensorimotor world, a naive agent can notice the existence of peculiar sensory changes which \emph{can be generated equivalently by sending a motor command (active mode) or by observing the consequences of environmental changes (passive mode)}. We later refer to these equivalent sensory changes as \emph{redundant sensory changes}.
For example the sensory change generated by the displacement of the agent with respect to an object can also be generated by an opposite displacement of the object with respect to the agent (see Fig.\ref{fig:redundancy}).
On the contrary sensory changes induced by a state change usually do not exhibit such a redundancy property. For instance the sensory change produced by an object changing color cannot be equivalently generated by the agent sending a motor command\footnote{This is true for the kind of motor capacities that biological and robotic systems usually have. The possibility of more exotic actions will be considered and discussed in Sec.~\ref{sec:Discussion}.}. Conversely the sensory change induced by the agent changing the sensitivity of one of its sensors cannot be equivalently generated by the environment.
The sensory redundancy that an agent can internally discover thus seems to correspond to what an external observer would characterize as the spatiality of changes in the agent-environment system.
Moreover this concept of redundancy naturally accounts for space being experienced as shared between the agent and the environment, as both of them play a symmetrical role in the sensory redundancy. In addition it accounts for space being subjectively content-independent since redundant sensory changes remain redundant independently of the objects present in the environment. Finally it also explains the subjective isotropy of space, as equivalent redundant sensory changes can be experienced regardless of the agent's and object's positions and orientations.

\begin{figure}[t]
\centering
\includegraphics[width=1\linewidth]{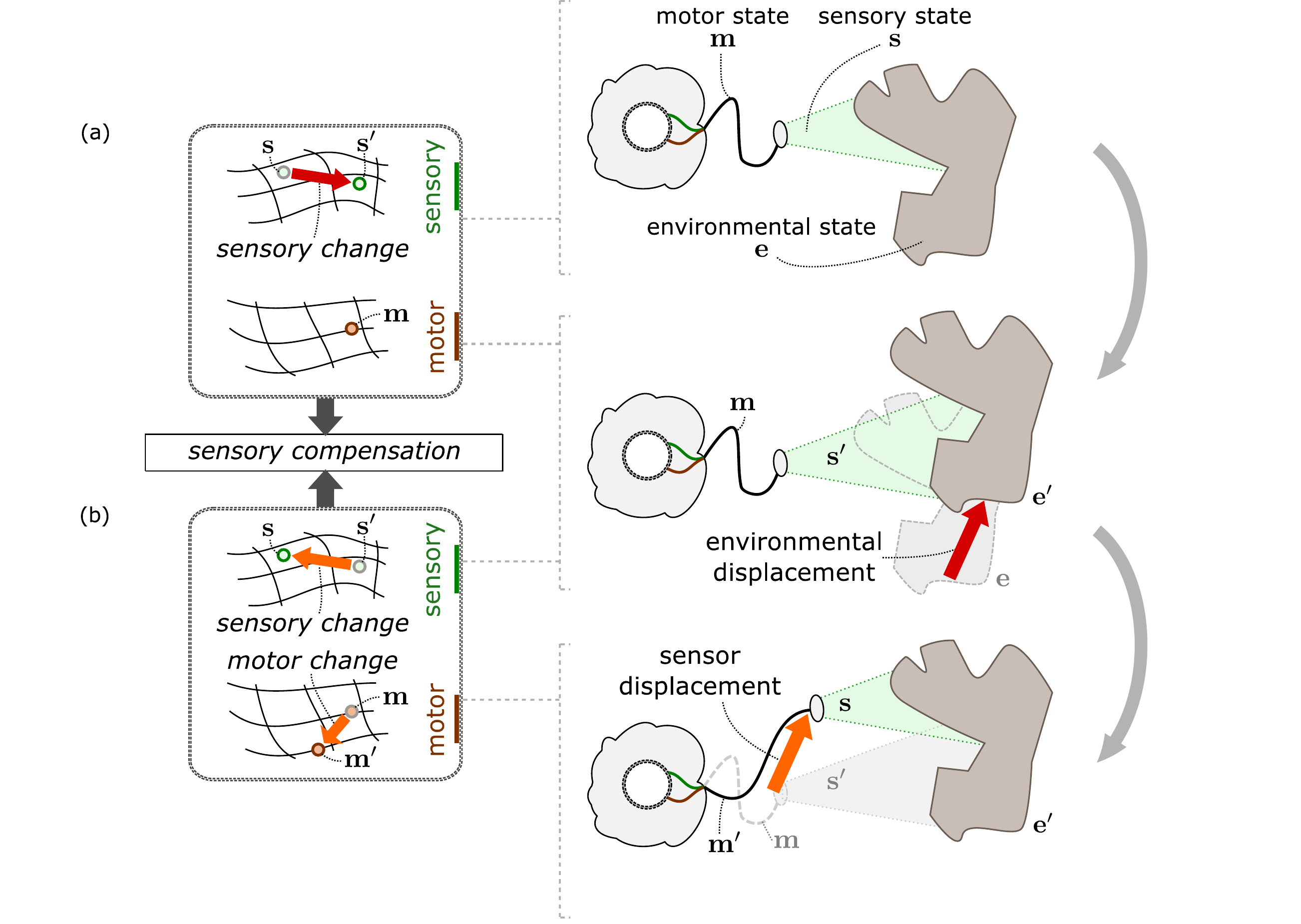}
\caption{Due to the sensory equivalence of displacements of the agent and of the environment, it is possible for one of them to \emph{compensate} for the other. By undergoing the same displacement as the environment (a), the agent can cancel out the resulting sensory change (b), and vice versa.}
\label{fig:compensability}
\end{figure}

The insight proposed by Poincar\'e was not focused on the concept of redundancy but on its corollary, the concept of \emph{compensability}.
Because there exists a set of redundant sensory variations that can be generated both by the agent or its environment, there is the possibility for one of them to compensate  for the (spatial) sensory variations of the other. 
As an example, the redundant sensory change produced by the displacement of an object in the environment can be compensated, or canceled out, by the same displacement of the agent, and vice versa (see Fig.\ref{fig:compensability}). As the relative positions of the agent and the object are identical before and after the two displacements, so is the agent's sensory input.
Interestingly this ability to compensate for spatial changes in the environment suggests that the agent could have an active role in the discovery of space. Instead of merely observing sensorimotor changes in the hope of detecting redundancy, the agent could be inclined to compensate them. A fundamental drive for such a behavior could be the necessity for the naive agent to control its sensorimotor experience, an endeavor in which discovering sensory changes that can be compensated at will is valuable. Space would thus correspond to the general structure underlying redundant/compensable sensory experiences.

\subsection{Formalization}
\label{sec:formalization}

In this section we propose a mathematical formalism to describe the sensorimotor experience of an agent and its relation with the environment. In particular the concept of \emph{point of view}, proposed in previous work \citep{Laflaquiere2015a}, is recalled and its limitations with regards to spatial knowledge are highlighted. Redundant sensory changes are then defined and their relation with compensable sensory changes made explicit.

\subsubsection{The sensorimotor experience:}
\label{sec:the sensorimotor experience}

In this work we consider so called "naive" agents which are connected to an unknown body and immersed in a world they have no a priori knowledge about. They only have access to raw sensory and motor information.
Although this distinction between the sensory and motor flows might seem artificial in a holistic description of the agent, it reflects the fundamental distinction between what the agent does not directly control (the sensory flow it receives) and what it does (the motor flow it generates).
At any moment we assume that the agent's sensory and motor states can be respectively described by vectors $\mathbf{s}$ and $\mathbf{m}$ in the corresponding linear spaces $\mathbf{S} = \mathbb{R}^R$ and $\mathbf{M} = \mathbb{R}^N$.
Although the nature and properties of the sensors and motors associated with this sensorimotor experience do not need to be specified, the sensorimotor experience is assumed to comply with some assumptions.
First we consider that the sensory state only carries exteroceptive information. Proprioceptive information is supposed redundant with the motor state and is ignored in the following developments.
Second we assume as a first approximation that the system's dynamic can be ignored. The agent's actuators are thus instantaneously controlled in position via the motor state, while its sensors generate the sensory state without any transient phase.

Note that this formalism does not exclude the possibility of a pre-processing of sensory inputs or post-processing of motor outputs. The sensory state $\mathbf{s}$ could potentially correspond to the output of some module(s) processing the raw inputs from the sensors. Similarly the motor state $\mathbf{m}$ could potentially feed some module(s) which generates the actual commands sent to the actuators.
If these modules cannot be directly controlled in any way by the agent, they are transparent to it and are considered part of its unknown body, or more globally part of the unknown world it interacts with.
As a result our approach does not require unprocessed data and is not in conflict with the fact that evolution most probably endowed our brains with hardwired pre- and post-processing modules. Furthermore our approach aspires to be code-independent. It should thus be robust to any data transformation that does not destroy the information the agent needs to capture.

The agent's sensorimotor experience does not provide a complete picture of the agent-environment system. We consider a subset of all possible environments which we assume can be parametrized by vectors $\mathbf{e}$ of a vector space $\mathbf{E} = \mathbb{R}^P$. This space of environmental states is of course different from the \emph{space} in which the agent-environment system is embedded and that we seek to characterize. For instance, if the only degree of freedom of the environment corresponds to an object swapping between 4 colors, then the environmental space $\mathbf{E}$ would be a discrete set of 4 states.
Although the naive agent does not have directly access to it, the environmental state $\mathbf{e}$ influences its sensorimotor experience. It shapes the mapping between the motor state $\mathbf{m}$ and the associated sensory state $\mathbf{s}$. Using the notation introduced in \citep{Philipona2003}, we denote $\phi$ this mapping parametrized by $\mathbf{e}$:
\begin{equation}
\mathbf{s}=\phi_{\mathbf{e}}(\mathbf{m}).
\label{eq:phi}
\end{equation}
This mapping captures the structure that the world induces in the sensorimotor experience. Along with the parameter $\mathbf{e}$, the mapping $\phi$ embodies everything it would be useful for the agent to know about the world, the properties of its environment, and the properties of its body. In particular the mapping must contain information about the existence and the structure of space.
Nevertheless, with only access to $\mathbf{m}$ and $\mathbf{s}$, the agent can only probe $\phi_{\mathbf{e}}$ via its sensorimotor flow. It has to ground properties of the world in its own sensorimotor experience.

\subsubsection{Points of view:}
\label{sec:points of view}

\begin{figure}[h]
\centering
\includegraphics[width=1\linewidth]{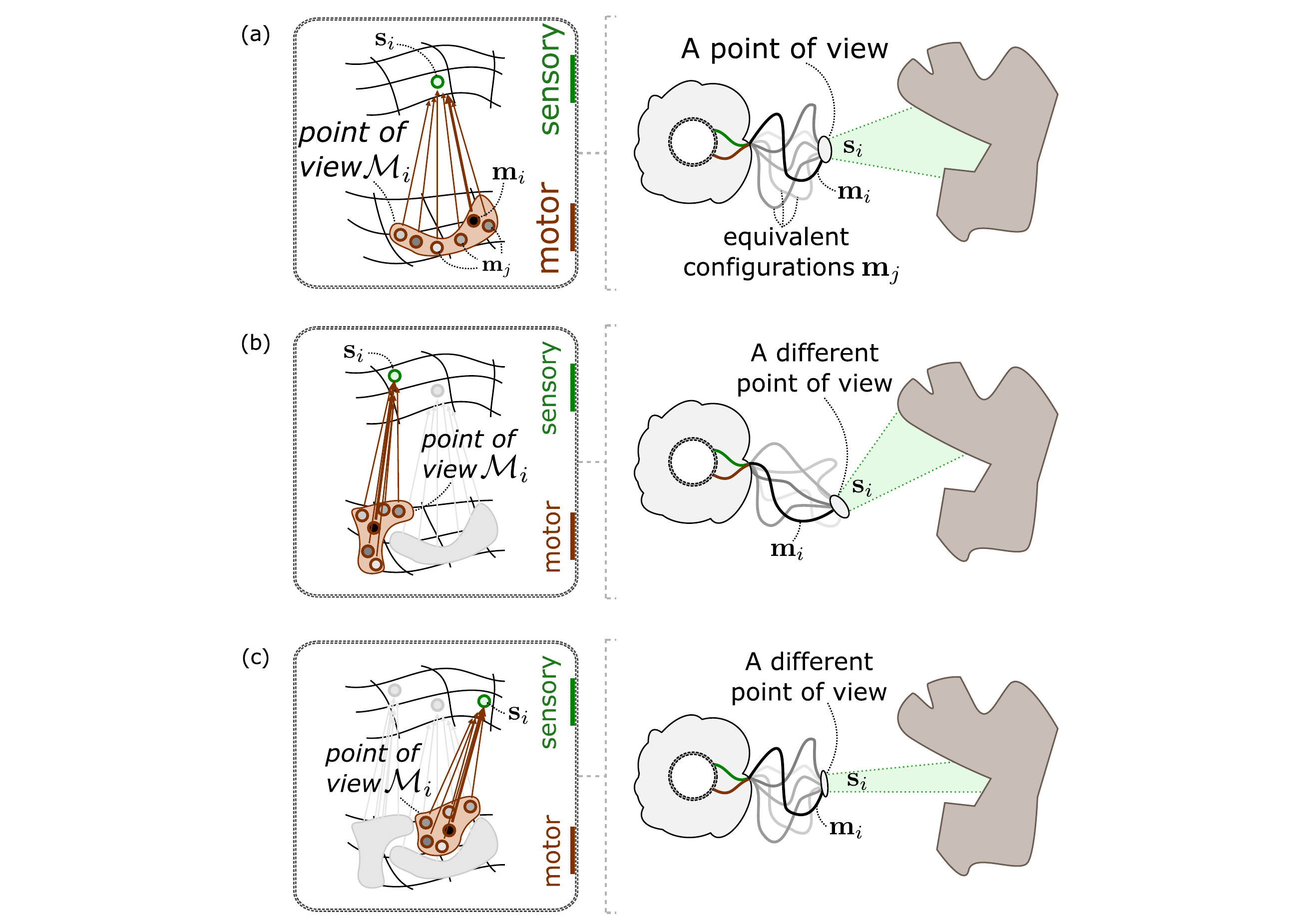}
\caption{(a) A \emph{point of view} corresponds externally to a configuration of the agent relative to its environment, and internally to a set of motor states that always generate equivalent sensory states. Nonetheless a point of view is not equivalent to a spatial configuration as it captures every degree of freedom of the agent, regardless of its spatial or non-spatial nature. Both a spatial change, as illustrated in (b), or a non-spatial change, like the change in sensor structure illustrated in (c), lead to a different point of view. In (c), the agent interprets its experience as a new point of view despite the fact that its sensor's spatial configuration has not changed compared to (a).}
\label{fig:pov_limits}
\end{figure}

We argue that the concept of space emerges through the discovery of specific sensorimotor invariants. In previous work the concept of \emph{point of view} was introduced as a proto-concept of spatial configuration by capturing a certain basic sensorimotor invariant \citep{Laflaquiere2015a}. This invariant corresponds to the sensory invariance induced by redundancy in the mechanical structure of the agent (see Fig.~\ref{fig:pov_limits}). Indeed different motor states can sometimes lead to the same effective configuration of the agent relative to the environment, and thus to the same sensory state. Moreover this equivalence of the motor states remains true regardless of the state of the environment.
Formally we associate to each motor state $\mathbf{m}_i$ a point of view, denoted $\mathcal{M}_i$, defined as the set of all motor states $\mathbf{m}_j$ that generate the same sensory states as $\mathbf{m}_i$ for any state of the environment $\mathbf{e}$:
\begin{equation}
\mathcal{M}_i =
\big\{
\mathbf{m}_j \text{ such that }
\phi_{\mathbf{e}}(\mathbf{m}_j) = \phi_{\mathbf{e}}(\mathbf{m}_i) 
\text{ for all } \mathbf{e}
\big\}.
\label{eq:equivalenceClass}
\end{equation}
Note that we assume here that the same motor states are possible for any state of the environment.
Following Eq.~\eqref{eq:equivalenceClass} we propose to extend the definition of the mapping $\phi$ so that it applies to points of view:
\begin{equation}
\mathbf{s}=\phi_{\mathbf{e}}(\mathcal{M}), \text{ where } \mathbf{s}=\phi_{\mathbf{e}}(\mathbf{m}) \text{ for all } \mathbf{m} \in \mathcal{M}.
\end{equation}
For an external observer, different sets $\mathcal{M}$ correspond to different configurations the agent's sensors can have relative to the environment, hence the name \textit{point of view} (see \citep{Laflaquiere2015a} for a more detailed discussion).
Nonetheless there exist exceptional situations in which the sensory equivalence captured by points of view could also be due to symmetries in the world or limits in the agent's sensory capabilities. An extreme example would be a half-empty world in which half of the motor states would be associated with the same sensory state. This peculiar property of the world would be invisible to the agent, since it would interpret it as just another particular point of view.

For the agent, identifying points of view is interesting on two levels. First it supports a more compact encoding of the agent's experience in which knowing the current point of view $\mathcal{M}_i$ instead of the particular motor state $\mathbf{m}_i$ is sufficient to describe the sensorimotor experience.
Second the knowledge of these sets has the advantage that it allows the agent to predict in advance the sensory state $\mathbf{s}$ associated with all motor states $\mathbf{m}_j \in \mathcal{M}_i$ from the experience in $\mathbf{m}_i$.
These two mechanisms fit well with the type of fundamental drives which have been typically proposed in the literature on unsupervised learning \citep{Bengio2012, Friston2010, Seth2014}.

As displayed in Eq.~\eqref{eq:equivalenceClass}, points of view have the interesting property of being \emph{independent from the state of the environment}, despite being derived from sensorimotor experiences which depend on it. Indeed $\mathbf{e}$ influences the sensory state $\mathbf{s}$ but not the motor space in which the sets $\mathcal{M}$ are defined. This property echoes the subjectively experienced \emph{content-independence} of space. It suggests that the notion of space should fundamentally be anchored in the agent's motor space.
Nonetheless points of view are not a direct internal representation of the agent's spatial configuration. Indeed they capture all the agent's effective degrees of freedom, which are not all necessarily spatial. As an example, imagine an agent which can control the spatial configuration of its sensors as well as some other non-spatial parameter of the sensors, like their sensitivity. Any change in this non-spatial parameter leads to a different sensory state and thus to a different point of view $\mathcal{M}$, when no spatial change actually occurred for the agent. An illustration of this limitation is proposed in Fig.~\ref{fig:pov_limits}(a)(c) with an agent that can change the strucutre of its sensor array.
The concept of point of view is thus insufficient to internally characterize the agent's spatial configuration independently from its other potential degrees of freedom. However we propose to use this compact representation of the motor experience in later developments to get rid of the agent's mechanical redundancy.

\subsubsection{Agent-environment redundancy:}
\label{sec:Agent-environment redundancy}

By observing the influence of the motor state on the sensory state, we have shown how a naive agent can detect a simple form of redundancy which can be captured in the form of points of view. Points of view internally represent the agent's external configuration, including both spatial and non-spatial parameters.
In order to isolate the spatial component of this configuration we claim that the agent  can detect a more sophisticated form of sensory redundancy. As described in \ref{sec:The sensorimotor imprint of space} it consists in a redundancy between how either the agent's motor state or the environmental state can change the agent's sensory state.
By identifying such redundant changes, the agent should be able to distinguish which points of view are related through spatial transformations and which are not.

Let us consider a reference state of the agent-environment system denoted $(\mathbf{e}_0,\mathbf{m}_0 \in \mathcal{M}_0)$. Given its dual nature, the system can undergo changes from two sources: the motor state and the environmental state.
In the first case we denote by $\mathbf{m}_a$ the motor state after a change $\mathbf{m}_0 \rightarrow \mathbf{m}_a$, and $\mathcal{M}_a$ its associated point of view after the corresponding change $\mathcal{M}_0 \rightarrow \mathcal{M}_a$.
For the sake of simplicity we drop hereinafter the notation based on the motor state in favor of the one based on the point of view, without any loss of generality (see Eq~\eqref{eq:equivalenceClass}).
In the second case we denote by $\mathbf{e}_b$ the environmental state after a change $\mathbf{e}_0 \rightarrow \mathbf{e}_b$.
According to Eq.~\eqref{eq:phi}, these state changes in the system generate sensory changes. Let $\mathbf{s}_{a}$ denote the sensory state associated with the configuration $(\mathbf{e}_0,\mathcal{M}_a)$ after the motor change:
\begin{equation}
\mathbf{s}_0 = \phi_{\mathbf{e}_0}(\mathcal{M}_0)
\rightarrow
\mathbf{s}_a = \phi_{\mathbf{e}_0}(\mathcal{M}_a).
\label{eq:phi_variationM}
\end{equation}
Similarly, let $\mathbf{s}_{b}$ denote the sensory state associated with the configuration $(\mathbf{e}_b,\mathcal{M}_0)$ after the environmental change:
\begin{equation}
\mathbf{s}_0 = \phi_{\mathbf{e}_0}(\mathcal{M}_0) \rightarrow \mathbf{s}_b = \phi_{\mathbf{e}_b}(\mathcal{M}_0).
\label{eq:phi_variationE}
\end{equation}
The redundancy we are interested in corresponds to sensory changes which can be generated both by a motor change or an environmental change (see Fig.~\ref{fig:redundancy}). Formally we're looking for states $\mathcal{M}_{a}$ and $\mathbf{e}_{b}$ such that:
\begin{equation}
\mathbf{s}_{a} = \mathbf{s}_{b}
\: \Longleftrightarrow \:
\phi_{\mathbf{e}_0}(\mathcal{M}_a) = \phi_{\mathbf{e}_b}(\mathcal{M}_0).
\end{equation}
%
According to the line of argumentation of section \ref{sec:The sensorimotor imprint of space}, this sensory equivalence should be due to the spatial nature of the changes $\mathcal{M}_0 \rightarrow \mathcal{M}_a$ and $\mathbf{e}_0 \rightarrow \mathbf{e}_b$. However it could also theoretically be due to some specific invariance in particular environments.
Indeed one could imagine a particular environment for which $\mathcal{M}_0 \rightarrow \mathcal{M}_a$ is not a spatial change but still has a sensory-wise equivalent $\mathbf{e}_0 \rightarrow \mathbf{e}_b$. As spatial changes, or more generally the structure of space, should be common to all possible environmental states, we rule out these outliers by looking for points of view $\mathcal{M}_a^*$ for which an equivalent environmental state exists for \emph{any} initial state $\mathbf{e}_0$ of the environment:
\begin{equation}
\text{For all } \mathbf{e}_0 \text{, there exists } \mathbf{e}_b
\text{ such that: }
 \phi_{\mathbf{e}_0}(\mathcal{M}^*_a) = \phi_{\mathbf{e}_b}(\mathcal{M}_0).
\label{eq:spatialchanges}
\end{equation}
The set of all $\mathcal{M}^*_a$ corresponds to all points of view related to $\mathcal{M}_0$ by a motor change which has a sensory-equivalent environmental change, regardless of the initial state of the environment. We denote it $\big\{ \mathcal{M}^* \big\}_0$ and claim that it internally represents all the agent's motor configuration which are related to $\mathcal{M}_0$ through spatial transformations.


\subsubsection{From redundancy to compensability:}
\label{sec:From redundancy to compensability}

As noted in section \ref{sec:The sensorimotor imprint of space}, H.Poincar\'e suggested that an naive agent can actively acquire spatial knowledge by discovering \emph{compensable} sensory changes. This concept of compensability is directly related to the one of redundancy introduced in the previous section.
{\color{black} Intuitively, it corresponds to the fact that if there exists a motor change which compensates for the sensory change induced by an environment change, then this motor change is itself redundant with the opposite environmental change. By compensating for sensory changes generated by the environment, the agent can thus discover motor changes which are redundant with environmental changes, i.e. displacements.
An illustration of the phenomenon is presented in Fig.~\ref{fig:compensability}.
}

More formally, let $(\mathbf{e}_0,\mathcal{M}_0)$ be the reference state of the agent-environment system, $(\mathbf{e}_b,\mathcal{M}_0)$ the state of the system after a change $\mathbf{e}_{0}\rightarrow\mathbf{e}_{b}$, and $\mathbf{s}_{b}$ its associated sensory state.
Let us now assume that the agent can find a point of view $\mathcal{M}_{a}$ which compensates for the experienced sensory change for any initial state of the environment $\mathbf{e}_0$, i.e.:
\begin{equation}
\phi_{\mathbf{e}_b}(\mathcal{M}_a) = \phi_{\mathbf{e}_0}(\mathcal{M}_0).
\label{eq:firstmove}
\end{equation}
According to Eq.~\eqref{eq:spatialchanges} and Eq.~\eqref{eq:firstmove} if we consider $\mathbf{e}_b$ as the initial state of the environment, making $(\mathbf{e}_b,\mathcal{M}_0)$ the new reference state of the system, then:
\begin{equation}
\mathcal{M}_a \in \big\{ \mathcal{M}^* \big\}_0 .
\end{equation}
This means that the change of point of view $\mathcal{M}_{0}\rightarrow\mathcal{M}_{a}$ compensates for the change $\mathbf{e}_{0}\rightarrow\mathbf{e}_{b}$ and is sensory-redundant with the opposite change $\mathbf{e}_{b}\rightarrow\mathbf{e}_{0}$ (see Fig.~\ref{fig:compensability}).
Consequently the agent can actively discover the spatially related points of view of $\big\{ \mathcal{M}^* \big\}_0$ by trying to compensate for the sensory changes its environment generates.\\

Before introducing the simulated system that will be used to illustrate the approach, let us summarize how space manifests itself in a naive agent's experience. While exploring the world, a naive agent can notice some simple invariants in its otherwise uninterpreted sensorimotor experience. For the sake of clarity we distinguished two kinds of such invariants, although they could both be captured by the agent at the same time during exploration. The first kind of invariant has been called point of view and corresponds
to the sets of motor states that are redundant in the sense that they provoke equivalent sensory states. These points of view internally characterize the agent's body configuration in the world, but capture both spatial and non-spatial aspects of this configuration. The second kind of invariant corresponds to the equivalence between certain sensory changes due to the environment (the agent does not act) and certain sensory changes due to the agent's motor changes. It can be seen as a redundancy not in the agent's body but between the agent's motor state and the state of the environment. An agent can actively discover this redundancy by compensating for the sensory changes it passively experiences. Using our usual spatial language, we refer to these redundant changes as displacements, and claim that they are the foundation of the agent's concept of space. This discovery of spatial knowledge could be based on a fundamental but simple drive for the agent to control its sensorimotor experience. Indeed it is very noteworthy and useful for an naive agent to discover invariants, which later allow it to predict the sensory outcome of some of its actions. Moreover, because these specific space-related invariants are independent of the content of space, they provide a useful generalization capacity to the agent.

\subsection{Simulated agent-object system}
\label{sec:Simulated agent-object system}

\begin{figure*}[t]
\centering
\includegraphics[width=0.9\linewidth]{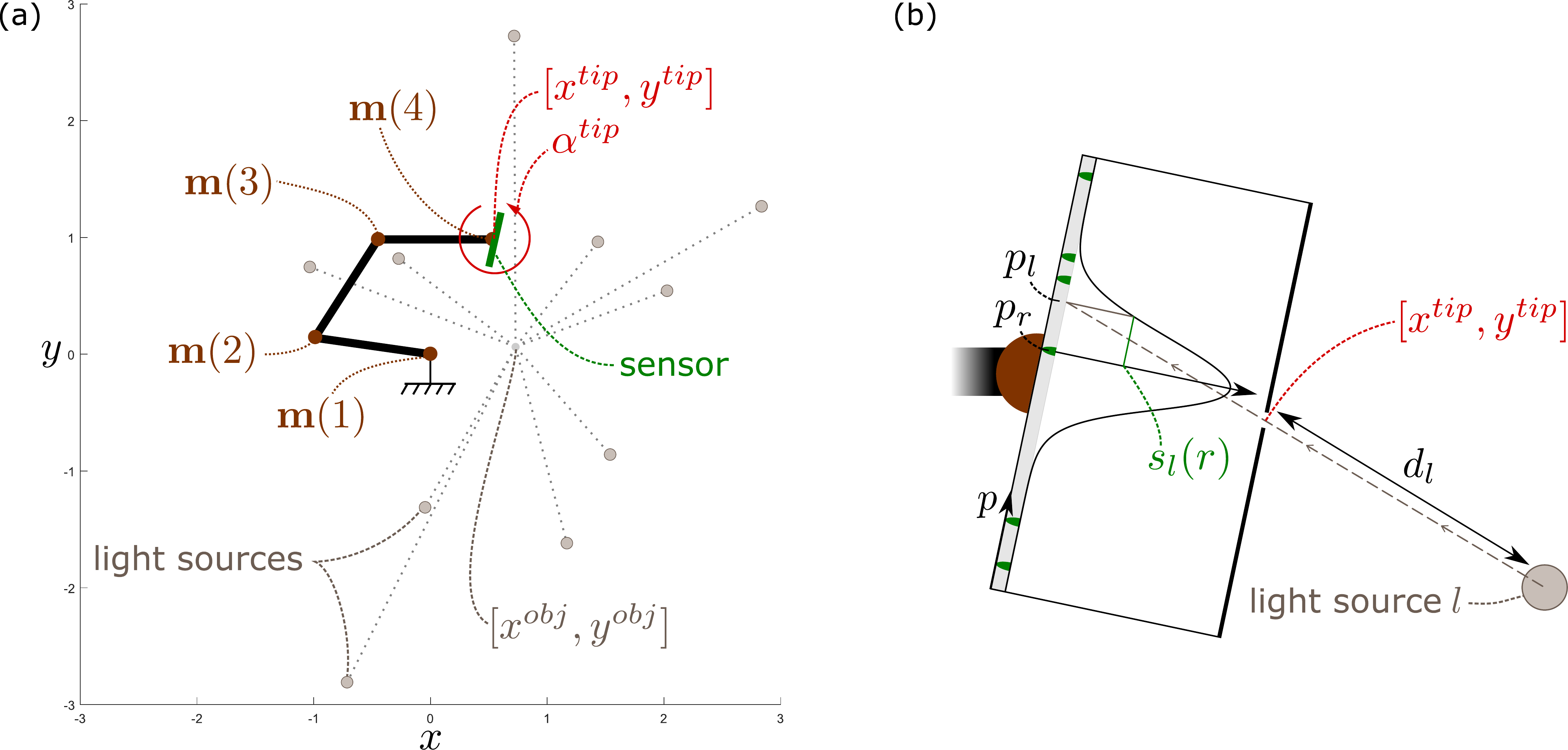}
\caption{Presentation of the simulated system. (a) The agent is a three-segment serial arm equipped with hinge joints, and an eye-like sensor that it can move in the plane. The environment consists in an object made up of $10$ point light sources that can translate rigidly in the plane. (b) Light emitted by the object is projected on the sensor's retina and generates sensations.}
\label{fig:system}
\end{figure*}

A simple simulated agent-environment system will be used in the following sections to illustrate how the topology and metric structure of space are accessible in a naive agent's sensorimotor experience.
The system, illustrated in Fig.\ref{fig:system}, consists in a robotic arm moving in the plane and observing an object. It is complex enough to avoid any trivial outcome of the simulation, while simple enough to ensure an intuitive analysis and visualization of the results.

\textit{The agent:}
The agent consists in a three-segment serial arm with a static base and a rotary sensor at its end-tip. Each of the three segments is of unitary length. The relative orientations in radians of its $K=4$ joints are controlled by the four elements $\mathbf{m}(k)$ of its motor state $\mathbf{m} = [\mathbf{m}(1),\mathbf{m}(2),\mathbf{m}(3),\mathbf{m}(4)]$. The arm tip has $3$ degrees of freedom in the plane: two for its position $[x^{tip},y^{tip}]$ and one for its orientation $\alpha^{tip}$. We call this triplet the \emph{spatial configuration} of the agent's sensor.
The end-sensor coarsely imitates an eye. It is made up of a pinhole lens in front of a linear retina covered with $6$ light-sensitive receptors. The position $p_r$ of each receptor ${r \in \{1,2,3,4,5,6\}}$ on the 1-D retina is randomly drawn at the beginning of the simulation. Each receptor is sensible to light in the environment. For a single point light source $l$ in the environment, the output $\mathbf{s}_l(r)$ generated by the receptor $r$ is defined as:
\begin{equation}
\mathbf{s}_l(r) = \frac{ e^{-(p_r-p_l)^2} } { d_l },
\end{equation}
where $p_l$ is the position of the light-source projection on the retina, and $d_l$ is the distance between the center of the lens and the light source $l$ (see Fig.~\ref{fig:system}).
For $L$ point light sources in the environment, the output $\mathbf{s}(r)$ of the receptor $r$ is the sum of the outputs for each source:
\begin{equation}
\mathbf{s}(r) = \sum_{l=1}^L \mathbf{s}_l(r).
\end{equation}
Note that given the projective nature of the sensor, any light source behind the lens is not projected on the retina and does not generate any excitation. Likewise, in this projective setting it is not relevant to specify the dimensions of the retina, which can consequently be considered unitary.
Finally the agent's sensory state is defined as the concatenation of all the receptor outputs: $\mathbf{s} = [\mathbf{s}(1),\dots,\mathbf{s}(6)]$.

It is important to highlight that the agent's structure, motors, and sensor have been defined arbitrarily. However our approach claims to be generic enough to lead to qualitatively similar results with a different design of the agent. The only design limitations are twofold. First the sensor should be rich enough so that two different configurations of the sensor generate two different sensory states. Second the agent should have a fixed base in the world. The latter assumption could however be loosened, as will be discussed in Sec.~\ref{sec:Discussion}.

\textit{The environment as an object:}
The environment is made up of $L=10$ point light sources randomly distributed around the agent at the beginning of the simulation. They are considered as part of a single rigid body, or "object", which can translate in the plane. The environmental state $\mathbf{e}$ is defined as the position $[x^{obj},y^{obj}]$ of the object's "center", which is arbitrarily set in $[0,0]$ at the beginning of the simulation.
Note that the object's center is defined only for the purpose of describing the simulation. It does not impact in any way the sensorimotor experience of the agent.
In the simulation, environmental changes $\mathbf{e}_i \rightarrow \mathbf{e}_j$ can be of two kinds. Either the object's center and all the light sources undergo a rigid translation in the plane, which corresponds to a spatial change $[\Delta x^{obj},\Delta y^{obj}]$. Or the lights sources undergo a random redistribution around the agent, which corresponds to a state change. During such a redistribution, the center of the object is considered unmoved. The state change can thus be seen as a change in the nature of the object but not in its position.

Note that the environment could be made more complex. Links between the different light sources could for instance be defined to explain the rigid structure of the object. However such an improvement would not alter the agent's sensorimotor experience if these links do not emit light.
Additionally note that the light sources have been randomly distributed and do not exhibit any specific structure. They only comply to the implicit constraint that the environment should present no symmetry from a sensory perspective. This way, two different configurations of the sensor relatively to the environment necessarily lead to two different sensory states.
Finally, although it was considered in \citep{Laflaquiere2015a}, we purposefully excluded the possibility of object rotations during spatial changes. There are two reasons for this choice. First, by artificially removing this spatial degree of freedom, we limit the spatial configuration to be discovered by the agent to the translations in the plane. This will ensure that we are able to visualize the internal representation built by the agent. Second, it indirectly means that rotations of the sensor will eventually be interpreted by the agent as non-spatial (state changes) because it cannot experience their compensability. This way, we can illustrate state changes of two different origins in the system: one from the environment and one from the agent itself.

In the following sections we apply the formalism developed so far to show how a naive agent can discover the topology of the spatial configuration of its sensor, as well as the regularity of the spatial metric.


\section{Capturing the topological structure of space}
\label{sec:Capturing the topological structure of space}

We claim that spatial knowledge is fundamentally based on the existence of sensory-redundancy between some motor changes of an agent and some changes of its environment.
In this section we illustrate, using the system introduced in \ref{sec:Simulated agent-object system}, how an agent can discover such a redundancy while interacting with its environment, and in particular by trying to compensate for sensory changes generated by the external world. The structure underlying the redundant experiences of the agent will be captured by building an internal representation whose topology will be shown to be identical to its sensor's spatial configuration.
First, we introduce a method to estimate the \textit{point of view} set associated with any motor configuration of the arm. Based on this proto-spatial concept we then show how the agent can identify which point of view changes are redundant (compensable) with environmental changes. Finally we propose a method to let the agent build an internal representation of its spatially related points of view.
Note once again that the focus of this paper is on identifying the sensorimotor structure underlying spatial knowledge and not on finding the best way to capture this structure. As a consequence the methods introduced below will sometimes rely on analytic methods to shortcut exploration phases which would otherwise require the development of complete machine learning solutions in themselves.

\subsection{Estimating the agent's points of view}
\label{sec:Estimating the agent's points of view}

The agent described in Sec.~\ref{sec:Simulated agent-object system} is mechanically redundant as it controls the 3 degrees of freedom of its sensor through 4 motorized hinge joints. This constitutes an interesting case study in which the motor state is not directly homeomorphic to the sensor's spatial configuration.
As introduced in \citep{Laflaquiere2015a} and described in Sec.~\ref{sec:points of view}, the agent's sensorimotor experience can be represented more compactly by defining \emph{points of view} $\mathcal{M}$ as sets of motor states $\mathbf{m}$ which generate identical sensory states $\mathbf{s}$ for any state of the environment $\mathbf{e}$ (see Eq.~\eqref{eq:equivalenceClass}).
Internally, considering points of view $\mathcal{M}$ instead of motor states $\mathbf{m}$ allows a more compact representation of the sensorimotor experience where equivalent motor states are reduced to a single state. It also saves replicating exploration of the motor states discovered in one point of view when subsequently exploring the environment. Externally this allows to describe the relevant parameters of the agent's configuration (spatial and non-spatial) while discarding mechanical redundancy of the body or other potential irrelevant symmetries in the system.
In this section we propose a method to estimate the point of view associated with any motor configuration of the arm.

\begin{figure*}[t!]
\centering
\includegraphics[width=1\linewidth]{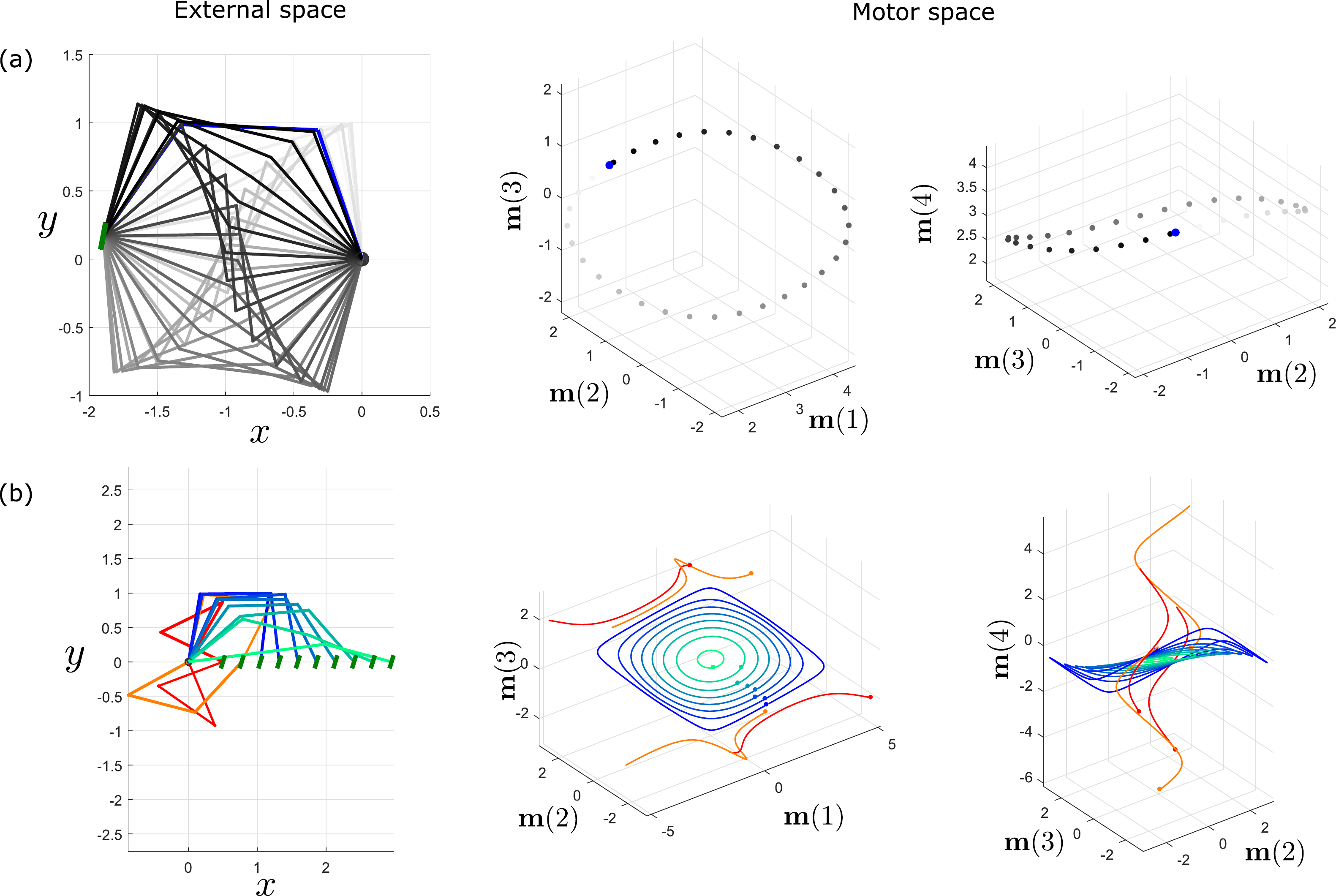}
\caption{(a) Visualization of the \textit{point of view} manifold associated with a random motor configuration (in blue). Configurations of the arm are shown on the left, and the corresponding motor states on the right. Moving along the looped 1D manifold in the motor space corresponds to a continuous change of the arm configuration which keeps the retina fixed.
(b) Visualization of the manifolds, on the right, associated with multiple arm configurations, on the left, such that the retina moves along the x-axis. When the arm's tip is more than 1 unit from its base, the points of view correspond to looped 1D manifolds. However when the tip is closer than 1 unit from the base, the manifold breaks into two disconnected submanifolds, due to the mechanical constraints of the arm (red and orange). An arm configuration belonging to both those submanifolds and producing the same spatial configuration of the sensor are shown on the left {\color{black}(in red and orange)}.}
\label{fig:singlekernel}
\end{figure*}

The system was designed to exhibit no peculiar symmetry with respect to the agent's sensory experience. Consequently the elements $\mathbf{m} \in \mathcal{M}$ of a point of view can be discovered by exploring a single environmental state $\mathbf{e}$. In a less favorable scenario, symmetries in the system for some environmental state $\mathbf{e}$ could result in different spatial configurations of the sensor generating the same sensory state. In such a case exploring multiple environments in which these symmetries are inconsistent would be required to sort out these "false positive" and correctly define each point of view $\mathcal{M}$. Nonetheless if every environmental state $\mathbf{e}$ consistently were to present the same symmetries, this property of the world would be directly captured in the sets $\mathcal{M}$ extended in this way without affecting the agent's ability to interact with its environment.

Because each set $\mathcal{M}$ is potentially infinite in the case of continuous motor commands, we propose to approximate it by a finite number $N=100$ of motor states $\mathbf{m} \in \mathcal{M}$.
The search for these samples could be performed by exhaustively (and/or randomly) exploring the motor space $\mathbf{M}$ and tracking which motor states generate the same sensory input. Other approaches could also be proposed like for instance the use of a neural network to directly build an internal space of points of view by capturing the topology of the sensory space(s) (see previous work \citep{Laflaquiere2013}).
Although these options are more realistic in terms of data accessible to the naive agent, we propose instead to use an analytic approach for the sake of computational efficiency.
It makes use of the arm's Jacobian and of its kernel to sample the subspace associated with $\mathcal{M}$. More precisely, given the mechanical structure of the arm, the hinge nature of its joints, and their continuous control law, each point of view $\mathcal{M}$ corresponds in the simulated system to a closed 1-D manifold in the 4-D motor space $\mathbf{M}$ (see Fig.~\ref{fig:singlekernel}(a)).
Starting from a motor state $\mathbf{m}_i$, we thus propose to sample states $\mathbf{m}_j \in \mathcal{M}_i$ by iteratively moving along this 1-D manifold with a small step $\varepsilon$ until the initial motor state is reached again. The direction of each step in the motor space is given by the kernel of the arm's Jacobian $\ker\big(J_f(\mathbf{m}_j)\big)$ which corresponds to a vector pointing in the direction along which local motor changes do not generate sensory changes:
\begin{equation}
\ker\big(J_f(\mathbf{m}_j)\big) = \ker \Bigg( \frac{\partial f}{\partial \mathbf{m}}\bigg|_{\mathbf{m}_j} \Bigg),
\end{equation}
where $f$ is the forward kinematic model:
\begin{equation}
f : \mathbf{m} \longmapsto (x^{tip},y^{tip},\alpha^{tip}).
\label{eq:forwardmodel}
\end{equation}
Starting from $\mathbf{m}_j = \mathbf{m}_i$, new samples are thus generated iteratively as follows:
\begin{equation}
\mathbf{m}_{j+1} = \mathbf{m}_{j} + \gamma\varepsilon \frac{\ker\big(J_f(\mathbf{m}_j)\big)}{\|\ker\big(J_f(\mathbf{m}_j)\big)\|},
\end{equation}
where $\varepsilon = 10^{-3}$ is the step size, and $\gamma =\pm 1$ is a parameter to ensure the successive steps consistently move in the same direction along the manifold (the kernel only defines the vector orientation).
Note that the vector added to $\mathbf{m}_{j}$ is never of zero magnitude as there is no degeneration in the arm's kinematics.
The sampling stops once at least $50$ motor states have been collected and the last sample $\mathbf{m}_j$ lies at a distance less than $10^{-2}$ from the initial motor state $\mathbf{m}_i$. Finally $N=100$ samples regularly distributed along the manifold are defined by interpolation over all the collected samples. These final samples are the only ones stored by the agent in order to limit memory usage during the simulation.
A more complete description of the sampling method is available in \citep{Laflaquiere2015a}.

Note that this sampling method induces a constraint on the range of points of view that can be estimated. Indeed it cannot sample sets $\mathcal{M}$ which are made of two disjoint subsets. This peculiar situation happens for any arm configuration such that the distance between the tip and the base is less than the length of its segments (see Fig.\ref{fig:singlekernel}(b)). Consequently these configurations are ignored during the simulation, which leaves a hole in the center of the arm's effective working space. This limitation is to be attributed to the analytic sampling method we proposed and not to the more general approach. Other methods could be implemented to discover the two disjoint subsets.

\subsection{Estimating spatially related points of view}
\label{sec:Estimating spatially related points of view}

The set of all points of view that an agent can estimate captures the variability of its sensor's external configuration. Yet not all changes between points of view are necessarily spatial. In this simulation we are interested in identifying the set $\big\{\mathcal{M}^*\big\}_0$ all points of view which are \emph{spatially} related to a configuration $\mathbf{m}_0 \in \mathcal{M}_0$ the agent is initially in. This means that we look for all points of view related to $\mathcal{M}_0$ through sensory changes redundant with environmental changes.

Given the continuous nature of the agent's motor experience, the set $\big\{\mathcal{M}^*\big\}_0$ is potentially infinite. We thus propose to approximate it by a finite number of points of view $\mathcal{M}_i$.
As described in \ref{sec:From redundancy to compensability}, the agent can actively identify points of view $\mathcal{M}_i \in \big\{\mathcal{M}^*\big\}_0$ by trying to compensate for sensory changes generated by the environment. We propose to take advantage of this mechanism by introducing a tracking-like behavior for the agent.
Starting from a random configuration of the system $(\mathbf{e}_0,\mathcal{M}_0)$, the exploration phase of the simulation consists in successively changing the environmental state and letting the agent try to compensate for the resulting sensory changes.
Each environmental change $\mathbf{e}_{i} \rightarrow \mathbf{e}_{i+1}$ is either a spatial change (translation of the object) or a non-spatial change (change of the object structure). In either case it generates a sensory change $\mathbf{s}_{i} \rightarrow \mathbf{s}_{i+1}$ that the agent then tries to compensate for. It does so by searching for a new configuration $\mathbf{m}_{i+1} \in\mathcal{M}_{i+1}$ such that the sensory variation is canceled out and the final sensory state is again $\mathbf{s}_{i}$ (see Fig.~\ref{fig:compensability}). Note that the environment is considered static during this search.
If the agent finds such a motor state $\mathbf{m}_{i+1}$, the associated point of view $\mathcal{M}_{i+1}$ is estimated (see Sec~\ref{sec:Estimating the agent's points of view}) and considered a sample of $\big\{\mathcal{M}^*\big\}_0$. If not, the experience is discarded, the arm resumes its previous motor configuration $\mathbf{m}_{i+1} = \mathbf{m}_{i}$, and waits for the next environmental change.
From an external perspective, the agent should be able to compensate for sensory changes which are generated by a spatial changes of the object. On the contrary, it should not be able to find a compensating motor configuration when the environmental change is non-spatial. Additionally any displacements of the object which would require  the sensor to move out of the arm's working space should also be discarded due to non-compensability.\\

\textit{Compensation:}
The search for compensating motor states $\mathbf{m}_{i+1}\in\mathcal{M}_{i+1}$ could be done by exhaustively exploring the motor space $\mathbf{M}$ or taking advantage of a tracking-like heuristic. Instead we propose an analytic approach for the sake of computational efficiency.
The method relies on the arm's kinematic model $f$ of Eq.~\eqref{eq:forwardmodel}, as well as the desired sensor configuration required to compensate for a displacement of the object's center $[\Delta x^{obj}, \Delta y^{obj}]$:
\begin{equation}
[x^{tip}_{i+1},y^{tip}_{i+1},\alpha^{tip}_{i+1}] = [x^{tip}_i,y^{tip}_i,\alpha^{tip}_i] + [\Delta x^{obj}, \Delta y^{obj}, 0].
\label{eq:displacementobject}
\end{equation}
The displacement of the object is null in the case of a non-spatial change.
The method consists in searching a minimum for the function
\begin{equation}
||\: f(\mathbf{m})-[x^{tip}_{i+1},y^{tip}_{i+1},\alpha^{tip}_{i+1}] \:||^2
\end{equation}
through a conventional minimum search method (Nelder-Mead simplex algorithm). If the minimized error is less than a threshold $\xi=10^{-3}$, the corresponding motor state
$\mathbf{m}_{i+1}$ is considered to compensate for the environmental change. Otherwise the sensory experience is discarded.\\

\begin{figure*}[t!]
\centering
\includegraphics[width=0.93\linewidth]{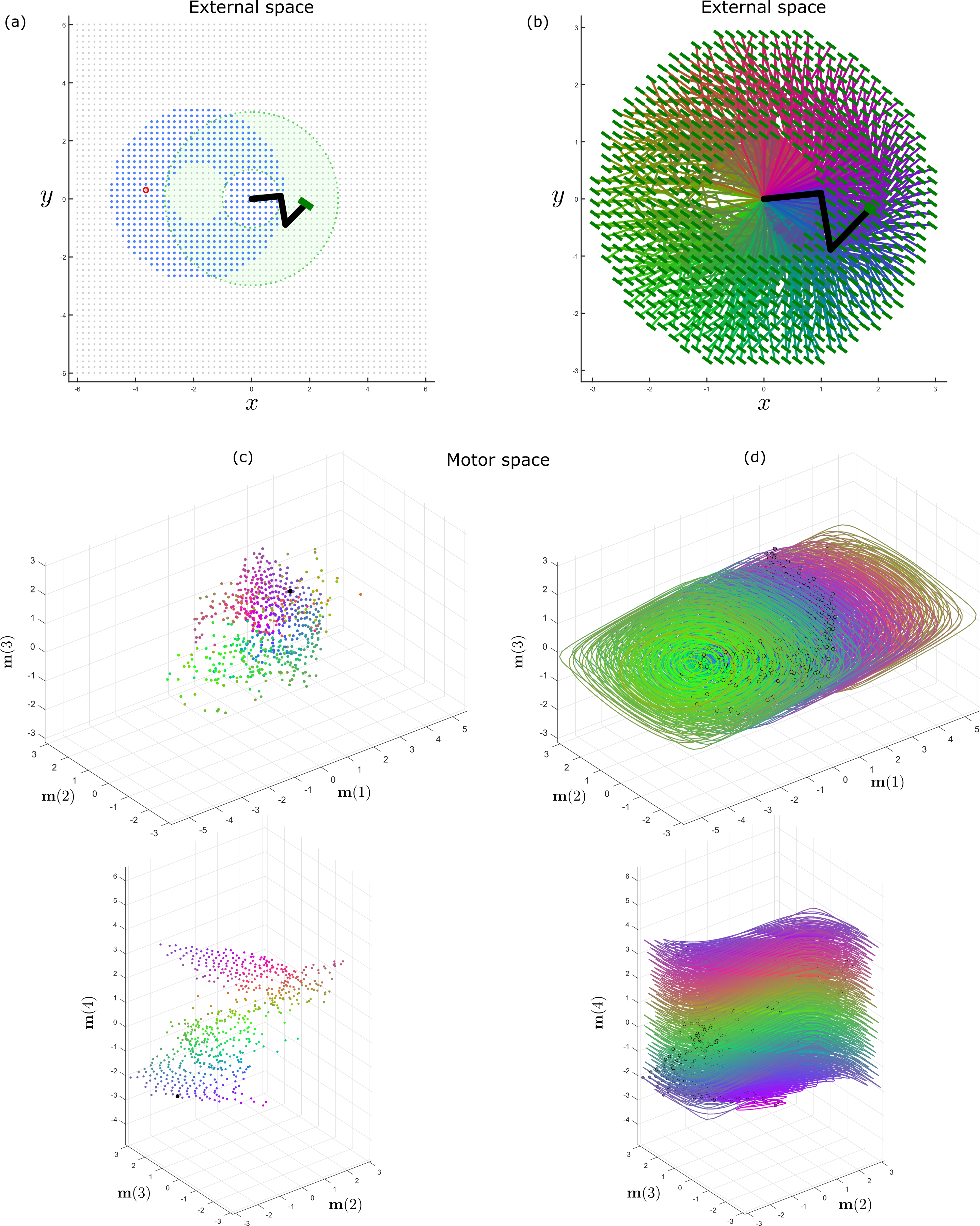}
\caption{(a) The initial arm configuration, and the grid of positions taken by the object during the simulation. The arm's working space is shown in green, and the positions of the object for which the agent was able to compensate the associated sensory change are displayed in blue. They are shifted to the left relatively to the working space because the initial position of the sensor lies on the right of the arm's base. One object position is outlined in red. It correspond to a theoretically compensable object displacement for which the search method did not converge.
(b) All the arm configurations found by compensating for the successive sensory variations generated by the object. The configurations are colored according to the angle the arm tip makes with the x-axis.
(c) All the motor states found by compensation and corresponding to the arm configurations of the previous panel. The initial motor state of the agent is displayed in bold black.
(d) All the estimated point of view sets associated with the compensating motor states of the previous panel.}
\label{fig:compensation_simu1}
\end{figure*}

\textit{Object displacements:}
During the simulation any translation of the object could be considered. However we propose to constrain them to a regular grid of positions in order to allow an intuitive interpretation of the simulation results. As illustrated in Fig.~\ref{fig:compensation_simu1}(a), the grid is made of $62^2$ regularly distributed positions centered on $[0,0]$. The width and height of the grid is set to 12 units, which ensures that the whole agent's working space can be covered by the compensatory arm displacements\footnote{The arm's maximum reach is 3 units long, which means that its working space has a maximum diameter of 6 units. This is the distance the grid must then cover in every direction.} regardless of its initial motor configuration $\mathbf{m}_0$.
During the exploration phase, the object is successively moved into each position in the grid, in no particular order. Between two such displacements it can also undergo a non-spatial change with a probability of 10\% by redistributing the light sources.\\

\textit{Results:}
Figure~\ref{fig:compensation_simu1} present the motor configurations $\mathbf{m}_i$ obtained by compensation of the object displacements and the corresponding external arm configurations for a given random initial motor configuration $\mathbf{m}_0 = [0.1, -1.5, 2.2, -3]$.
Over the $64^2$ positions taken by the object during the exploration phase, $648$ led to a compensation experience. The others were discarded because the search method was unable to converge to a satisfying solution. 
The found motor configurations are such that the end-tip sensor regularly covers the planar working space of the agent while maintaining a constant orientation.
This seems to indicate that the agent successfully identified translations of its sensor as spatial changes, while rotations of its sensor are not considered spatial due to a deliberate lack of compensability in the simulation.
Note that a position lying inside the working space has also been falsely identified as non-spatial due to a non-convergence of the search method (see red outlined position in Fig.~\ref{fig:compensation_simu1}(a)).\\
Finally Fig.~\ref{fig:compensation_simu1} also presents the estimated points of view sets $\mathcal{M}_i \in \big\{\mathcal{M}^*\big\}_0$ associated with the motor states $\mathbf{m}_i$ discovered by compensation during the exploration. These estimated manifolds represent the internal structure that was actually captured by the agent in its motor space.

\subsection{Building a topological representation of spatial configuration}
\label{sec:Building a topological representation of spatial configuration}

As can be observed in Fig~\ref{fig:compensation_simu1}(d), the spatially related manifolds $\mathcal{M}_i \in \big\{\mathcal{M}^*\big\}_0$ seem to exhibit an underlying topological structure in the motor space. Given the continuous nature of the arm's forward model, we indeed expect the set $\big\{\mathcal{M}^*\big\}_0$ to define a manifold. Moreover we expect this manifold to be homeomorphic to the space of the sensor's position $[x^{tip},y^{tip}]$.
We propose to capture its topology in a low-dimensional representation of the points of view $\mathcal{M}_i \in \big\{\mathcal{M}^*\big\}_0$.
In contrast with previous work where the topology of the set of \emph{all} points of view was estimated \citep{Laflaquiere2015a}, we aim here at capturing the topology of spatially related points of view only. The final internal representation built by the agent should thus be a representation of the sensor's spatial configuration.

A simple way to define a topology is to introduce a metric on the considered set $\big\{\mathcal{M}^*\big\}_0$. This metric will also be used to project the data into a low-dimension representational space .
%
We propose to use the Hausdorff metric to define a distance $d_{\mathcal{M}}(.,.)$ for each pair in $\big\{\mathcal{M}^*\big\}_0$.
The Hausdorff distance is the greatest of all the distances from a point in one set to the closest point in the other set.
The distance between two points of view is thus defined as follows:
\begin{equation}
\begin{aligned}
d_{\mathcal{M}}(\mathcal{M}_i,\mathcal{M}_j) = \max\Big\{\: \sup_{\mathbf{m}_a\in \mathcal{M}_i} \inf_{\mathbf{m}_b\in \mathcal{M}_j} d_\mathbf{m}(\mathbf{m}_a,\mathbf{m}_b) \:,\: \\ \sup_{\mathbf{m}_b\in \mathcal{M}_j} \inf_{\mathbf{m}_a\in \mathcal{M}_i} d_\mathbf{m}(\mathbf{m}_a,\mathbf{m}_b) \:\Big\},
\label{eq:hausdorff}
\end{aligned}
\end{equation}
with $d_\mathbf{m}$ the metric defined in the motor space $\mathbf{M}$. The latter is the natural Euclidean metric but slightly modified to take into account the periodicity induced by the arm's hinge joints. Indeed the physical structure of the arm implies that the agent's experience is invariant modulo $2\pi$ on any of its motor. The metric between two motor states in the motor space is thus defined as follows:
\begin{equation}
d_\mathbf{m}(\mathbf{m}_a,\mathbf{m}_b) = \sqrt{ \sum_{k=1}^K \arccos \Big( \cos \big( \mathbf{m}_a(k) -\mathbf{m}_b(k) \big) \Big)^2 } ,
\end{equation}
where $\mathbf{m}(k)$ is the $k$-th element of the vector $\mathbf{m}$.
This motor metric ensures that the distance between two elements $\mathbf{m}(k)$ of two motor states is not greater than $\pi$, which means that the distance $d_\mathbf{m}(\mathbf{m}_a,\mathbf{m}_b)$ between two motor states is not greater than $2\pi$ for $K=4$.
Although we impose this periodicity property in the motor metric, the agent could also potentially discover it, since the periodicity induced by the arm's hinge joints implies an equivalent sensory periodicity.
Finally note that the metric we define on the set $\big\{\mathcal{M}^*\big\}_0$ is derived from the motor space and has nothing to do with the external metric which applies to the sensor's spatial configuration.

The metric $d_\mathcal{M}$ is applied to compute distances between all pairs in $\big\{\mathcal{M}^*\big\}_0$. These distances enable the use of a dimension reduction method to build a low-dimensional representation of the set.
We chose the Curvilinear Component Analysis (CCA) method to perform the projection \citep{Demartines1997}. Like its famous couterpart Isomap \citep{Tenenbaum2000}, CCA is a non-linear projection method, except that it does not rely on the assumption that the manifold underlying the data is developable \citep{Lee2007}. To preserve the data topology during the projection, CCA preserves small pairwise distances while allowing long range distortions (unfolding).
For the sake of visualization we arbitrarily project the data into two separate 2D and 3D representational spaces. This manual tuning of CCA's output dimension could potentially be replaced by a data-driven estimation of the manifold intrinsic dimension \citep{Laflaquiere2012}.\\

\begin{figure}[t!]
\centering
\includegraphics[width=0.9\linewidth]{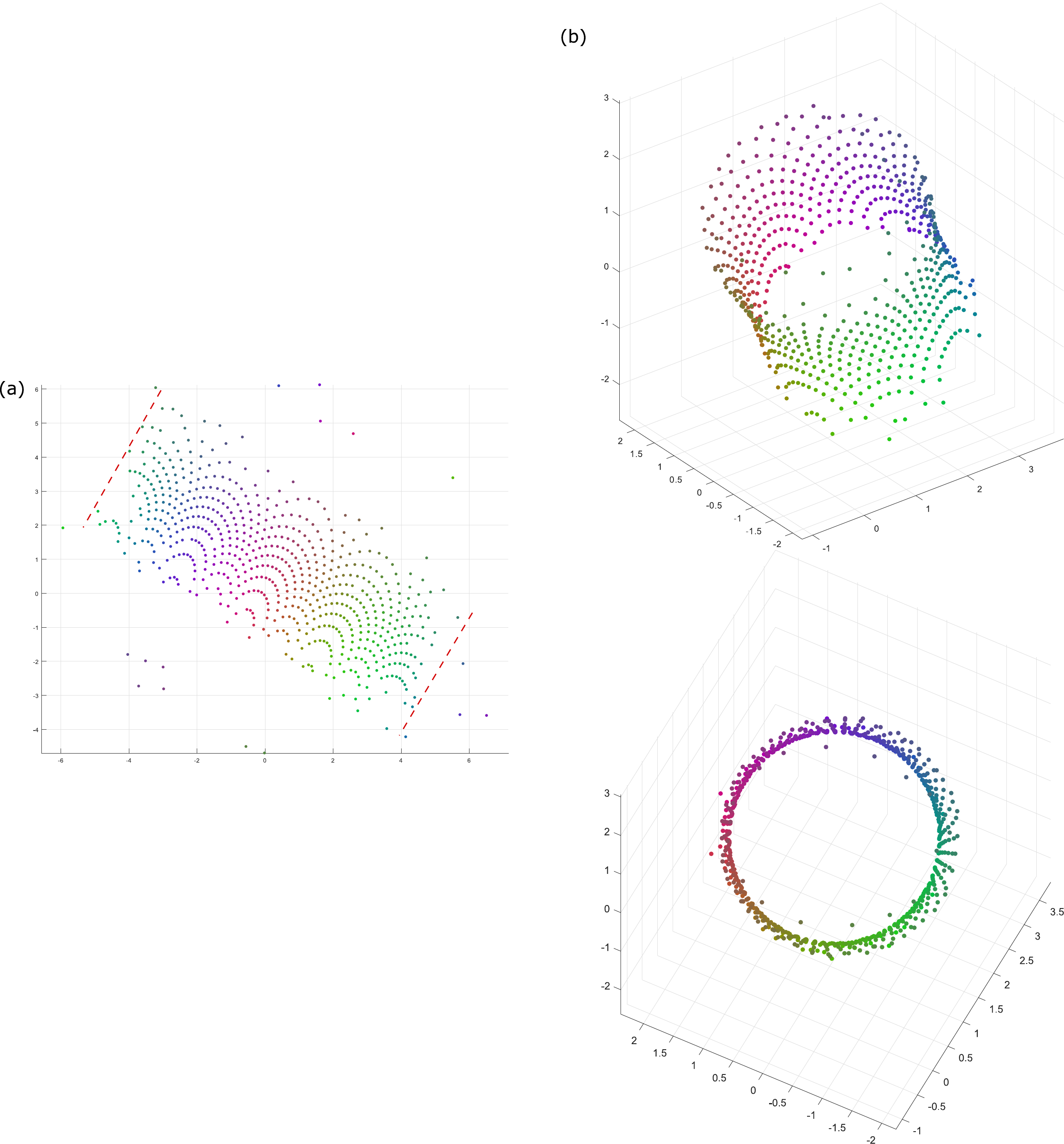} 
\caption{Projections of the points of view $\mathcal{M} \in \big\{\mathcal{M}^*\big\}_0$ in 2D (a) and 3D (b), using the same color code as in Fig.\ref{fig:compensation_simu1}. The underlying manifold of points of view appears to have a 2D cylindrical structure. Its curvature is too strong for CCA to project it in 2D without a cut (red dashed lines).}
\label{fig:projections}
\end{figure}

\textit{Results:}
Figure~\ref{fig:projections} presents the projection of the set $\big\{\mathcal{M}^*\big\}_0$ in 2D and 3D.
It appears that the manifold underlying the data has an intrinsic dimension of 2. Nonetheless it is highly curved, as revealed by its 3D projection, and CCA has been unable to unfold it in 2D without a cut (see Fig.~\ref{fig:projections}(a)).
This result is however consistent with the experiences collected by the agent during the exploration phase. It did indeed translate its sensor in a plane to compensate for the object's displacements, which effectively corresponds to a two-dimensional manifold. Moreover the consistent color code used in the figures show that the topology of this external plane has been correctly captured by the agent. The whole agent's working space and the neighborhood relations of the sensor spatial configurations are correctly represented (this is true for the 3D projection, and for the 2D projection if we omit the cut induced by CCA to flatten the manifold).
The internal representation built by the agent based on its own sensorimotor experience is thus a suitable topological representation of the spatial configuration of its sensor.
An element-wise comparison of the distances between the points of view in the external Euclidean metric and the Hausdorff metric derived from the motor space is proposed in Fig.\ref{fig:comparedmetrics}(a). The same comparison between the Hausdorff metric and the low-dimensional metric after projection in 2D/3D is also presented in Fig.\ref{fig:comparedmetrics}. They show that the metric estimated by the agent differs from the external metric and that the low-dimensional projections further distort it (see Fig.\ref{fig:comparedmetrics}(b)).

\section{Capturing the regularity of the spatial metric}
\label{sec:Capturing the regularity of spatial metric}

\begin{figure*}[t!]
\centering
\includegraphics[width=1\linewidth]{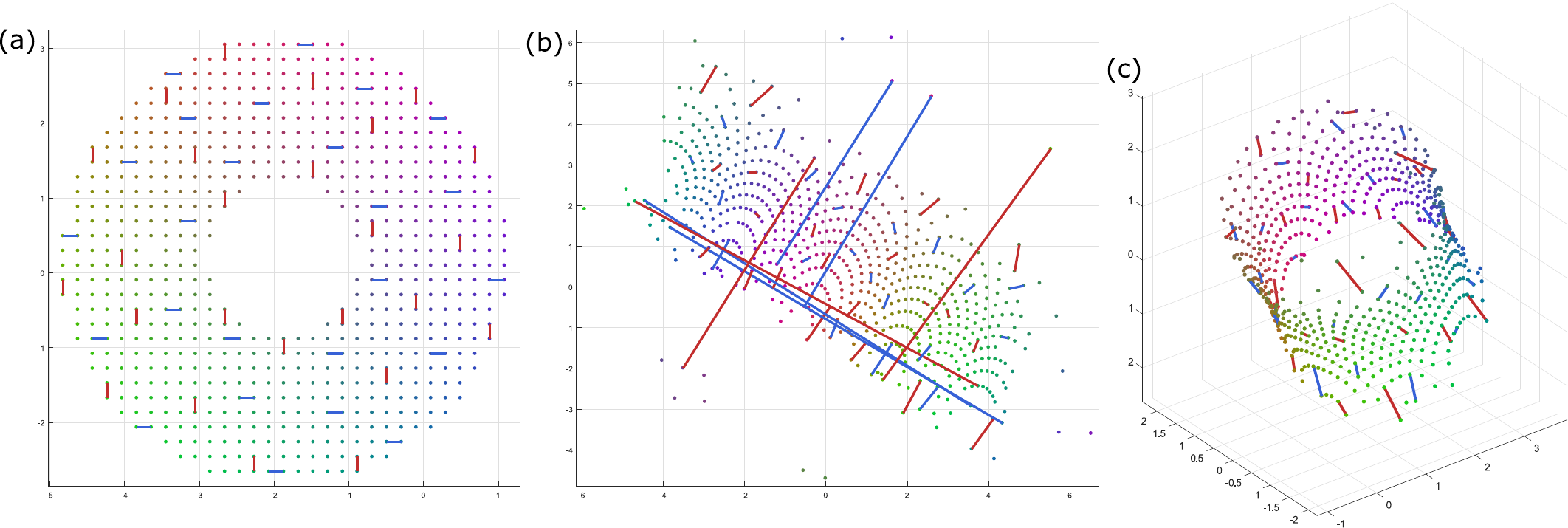}
\caption{Illustration of the metric difference between the external grid of sensor positions (a), and its internal representations in 2D (b) and 3D (c). All red (resp. blue) transitions in the grid correspond to equivalent external vertical (resp. horizontal) displacements, and thus have the same magnitude. However they are internally represented by vectors or various magnitude. Note that abnormally large vectors in (b) are due to the manifold being cut during the projection in 2D.}
\label{fig:metricdifferences}
\end{figure*}

By experiencing compensable sensory changes, a naive agent can discover motor changes which are redundant with environmental changes from a sensory perspective. These noteworthy experiences exhibit an underlying structure homeomorphic to the agent's spatial configurations. As presented in previous section in the case of a simple simulated arm, this structure can be captured explicitly in an internal representation whose topology is identical to the one of the external space the agent is moving in.
Yet the metric of the internal representation noticeably differs from the one of the external space, as revealed by the shape of the projected manifold in Fig.~\ref{fig:projections}. The data cloud appears there as a tube instead of a flat surface that would correspond to the shape of the actual grid of positions the sensor has been in during the simulation (see Fig.~\ref{fig:compensation_simu1}(a)).
This difference is expected as the agent does not have direct access to the external Euclidean metric. The internal representation is instead derived from the motor metric. This results in two equivalent displacements in the external space to be represented internally by two vectors of different magnitude and orientation, as illustrated in Fig.~\ref{fig:metricdifferences}.

However we claim that a naive agent can discover the metric regularity of the external space and modify its internal metric accordingly. This can be done through the discovery of new invariants accessible through a more sophisticated exploration of the environment.
In this section we describe this exploration strategy and how the metric regularity manifests itself in the agent's sensorimotor experience using the same simulated system as before. We then propose a method to modify the internal metric estimated by the agent and consequently its whole internal representation. Finally we analyze the resulting final representation of the sensor's spatial configuration and evaluate it on a simple reaching task.

\subsection{Discovering the spatial metric regularity}
\label{sec:Discovering the spatial metric regularity}

We subjectively experience space as homogeneous or isotropic: the length of an object does not change depending on the point of view of the observer. For instance the distance between the two ends of a rigid stick is constant, regardless of where it is positioned in our field of view. This metric invariance is of course not to be mistaken with the stick \emph{appearance} which does vary depending on our point of view (for example it grows smaller with distance).
This metric regularity is so far not captured in the internal representation built by the agent. Let us imagine a stick of length equal to the distance between two positions in the external grid of object positions.
Moving from one end of the stick to the other always requires a displacement of the same magnitude, regardless of where the stick is placed in the working space. On the contrary, the projection of these displacements in the internal representation leads to changes of different length depending on the position of the stick (see Fig.~\ref{fig:metricdifferences}).

\begin{figure}[t!]
\centering
\includegraphics[width=1\linewidth]{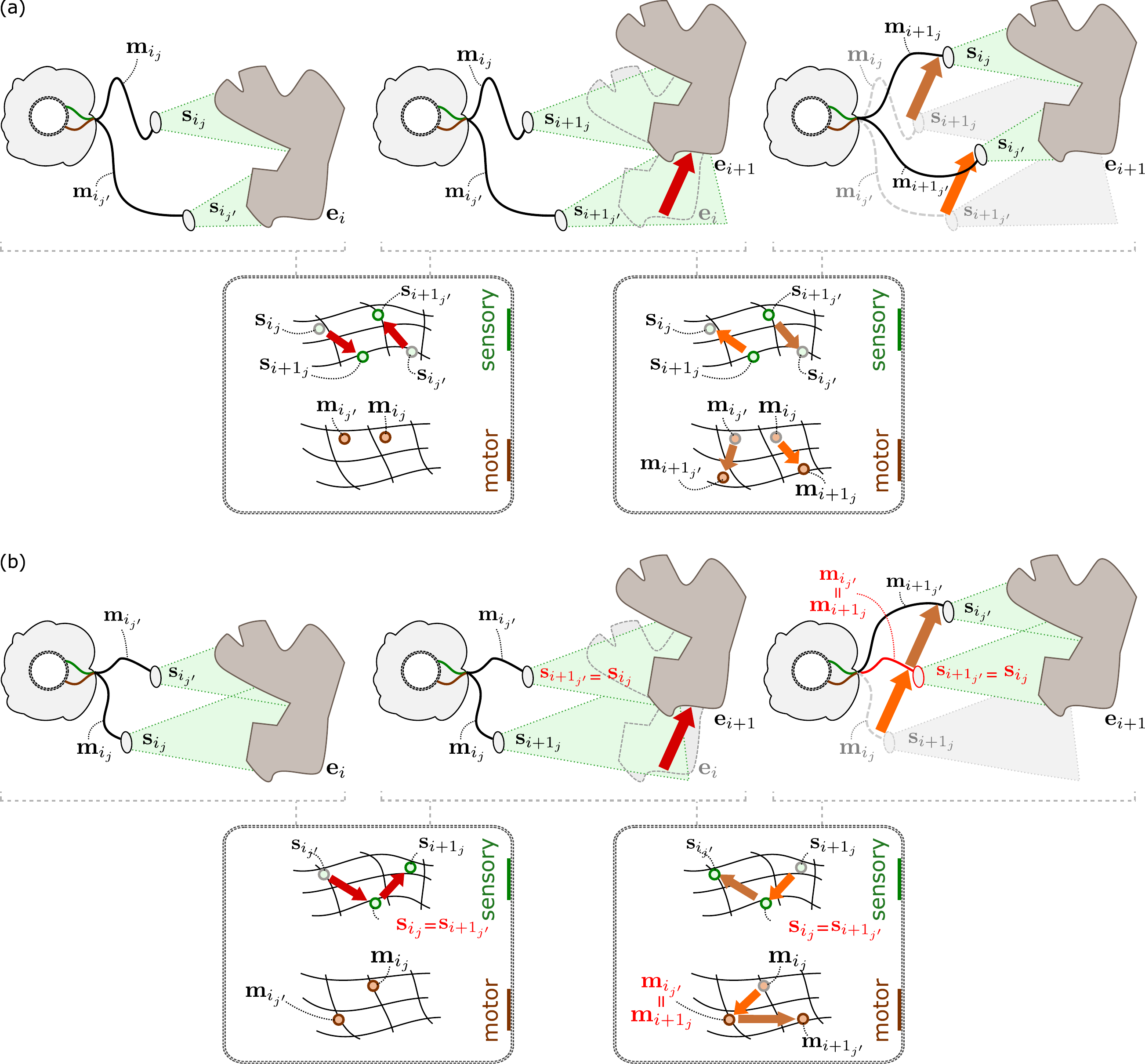}
\caption{(a) {\color{black} A single environmental change (here $\mathbf{e}_{i} \rightarrow \mathbf{e}_{i+1}$) can be compensated from multiple initial motor states (here $\mathbf{m}_{i_j}$ and $\mathbf{m}_{i_{j'}}$). This way, the agent can infer that the different corresponding motor changes (here $\mathbf{m}_{i_j} \rightarrow \mathbf{m}_{i+1_j}$ and $\mathbf{m}_{i_{j'}} \rightarrow \mathbf{m}_{i+1_{j'}}$) compensate for the same external displacement, and should thus be of equal magnitude.}
(b) If the environmental displacement corresponds to the displacement between the initial motor states, then some new motor states found by compensation actually correspond to some initial motor states already stored in memory.
{\color{black}Here the motor state $\mathbf{m}_{i+1_{j}}$ found by compensation corresponds to the initial motor state $\mathbf{m}_{i_{j'}}$}
(see the arm configuration in red).
{\color{black} Motor states $\mathbf{m}$ have been used in this illustration for the sake of simplicity, but the same rationale can be applied to points of view $\mathcal{M}$.}
}
\label{fig:regularization}
\end{figure}

In order to capture the external metric regularity, the agent has to discover that different point of view changes $\mathcal{M}_{i} \rightarrow \mathcal{M}_{i'}$ correspond to equivalent external displacements. This can be achieved by building on the already acquired sensorimotor knowledge and performing an exploration inspired by the invariance that was just illustrated with the stick example.
However the nature and length of a specific object, like a stick, is not an information directly accessible to the agent. Instead we thus consider displacements of objects as a "measuring rod" in the external space.
The rationale behind this choice is once again that an agent should discover the properties of space through the displacements it allows. This way compensability can be used to characterize these sensorimotor experiences and modify the agent's internal representation accordingly, regardless of the specific content of space.

We propose a more sophisticated way for the agent to explore its environment. It is similar to the one introduced in the previous simulation (see Sec.~\ref{sec:Estimating spatially related points of view}) except that multiple points of view $\mathcal{M}_{i_j}$ are considered at each step of the simulation instead of a single $\mathcal{M}_i$.
Its principle is illustrated in Fig.~\ref{fig:regularization}(a).
Initially the agent observes an object, in the envionmental state $\mathbf{e}_0$, from different points of view denoted $\mathcal{M}_{0_j}$, respectively generating sensory states $\mathbf{s}_{0_j}$, where index $j \in \{1,\dots,J\}$ denotes the point of view.
Similarly to the previous simulation, we then let the environmental state iteratively change $\mathbf{e}_{i} \rightarrow \mathbf{e}_{i+1}$ via either spatial or non-spatial transformations. This results in sensory changes $\mathbf{s}_{i_j} \rightarrow \mathbf{s}_{i+1_j}$ associated with each point of view $j$. The agent can try to compensate each of them by finding a point of view $\mathcal{M}_{i+1_j}$ which cancels out the sensory change and generates $\mathbf{s}_{i_j}$ again.
For each point of view $j$, if the environmental change is compensable, the agent can find such a suitable $\mathcal{M}_{i+1_j}$.
Interestingly all the internal changes $\mathcal{M}_{i_j} \rightarrow \mathcal{M}_{i+1_j}$ correspond to a single change in the environment. This is equivalent to exploring the same stick from different starting points of view. They should thus all have the same magnitude in the internal metric and comply to the following constraint:
\begin{equation}
\forall (j,j') \in \{1,\dots,J\} \:,\:  \widehat{d}_{\mathcal{M}}(\mathcal{M}_{i_j}, \mathcal{M}_{i+1_j}) = \widehat{d}_{\mathcal{M}}(\mathcal{M}_{i_{j'}}, \mathcal{M}_{i+1_{j'}}).
\end{equation}
where $\widehat{d}$ denotes the desired modified metric.
From an external perspective, the exploration phase would look like the agent first observes the environment from different points of view, then experiences the environment moving, and tries to find new motor configurations which ensure the same set of relative positions with the environment. It is analogous to a multi-point-of-view tracking behavior. Note that we assume here that the environment does not move while the agent searches for the multiple compensating points of view.
When the environmental change is a state change, the agent cannot find suitable points of view $\mathcal{M}_{i+1_j}$ to compensate for the associated sensory change. In this case the experience is discarded, the agent resumes its previous configurations $\mathcal{M}_{i_j}$ and waits for the next environmental change.
For some $j$, non-compensability can also occur for some spatial environmental changes which would require the sensor to move out of the agent's working space. In those cases, the experience is discarded like in the case of state changes.\\

\begin{figure*}[t!]
\centering
\includegraphics[width=0.82\linewidth]{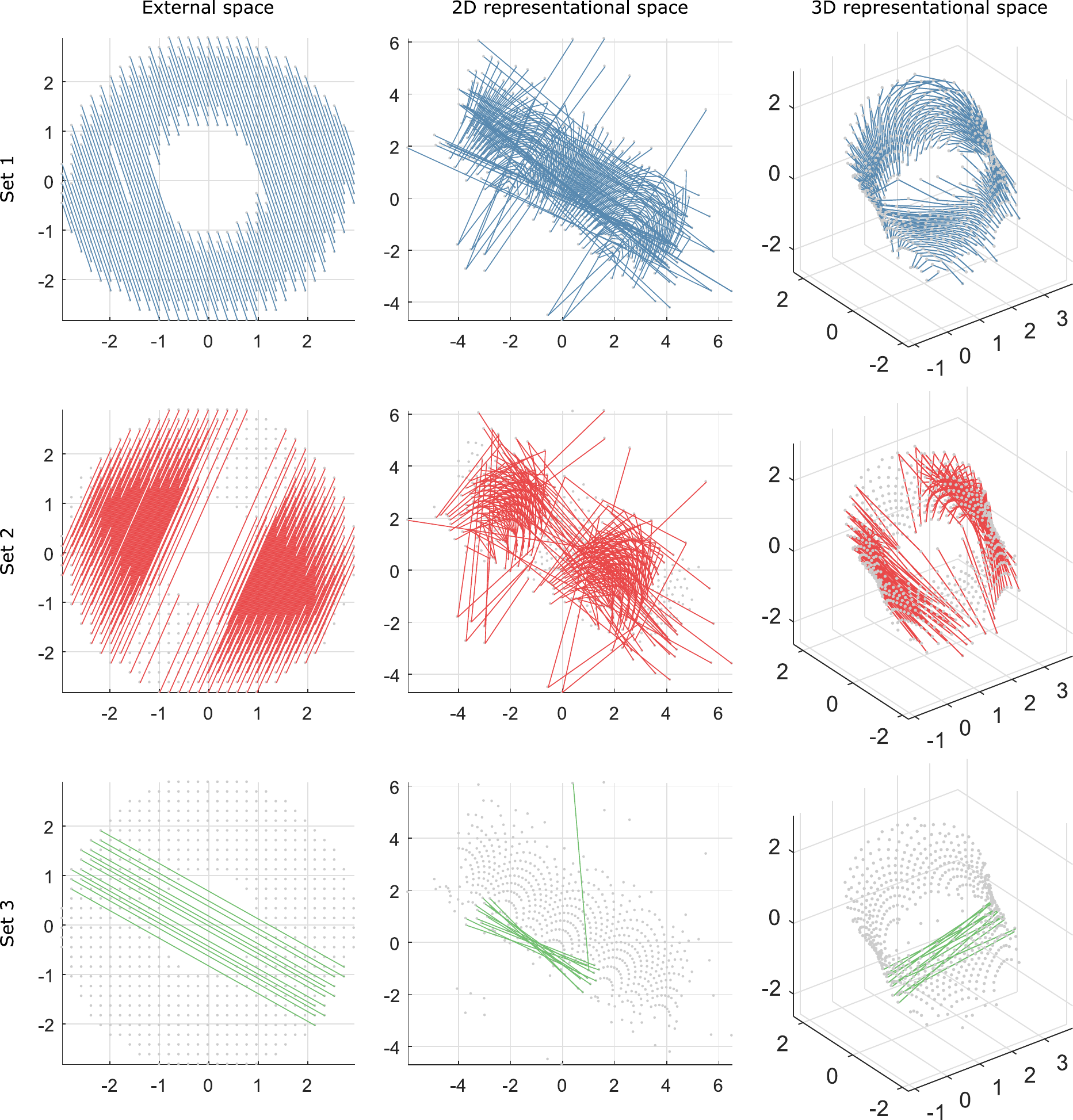}
\caption{Three examples of sets of point of view changes $\mathcal{M}_{i_j} \rightarrow \mathcal{M}_{i+1_j}$ associated with three environmental displacements. They are shows in the external space, where they correspond to identical displacements, and in the 2D and 3D representational spaces, where they correspond to changes of various magnitudes.}
\label{fig:equalitysets}
\end{figure*}

\textit{Optimized simulation setting:}
The exploration strategy we just proposed has been described in a generic form. For instance the different initial points of view $\mathcal{M}_{0_j}$ can be taken anywhere in the space of points of view. This means that they are not necessarily related through spatial changes and do not necessarily belong to a unique set $\big\{\mathcal{M}^*\big\}$. In the simulated system where rotations are considered state changes, this would correspond to points of view $\mathcal{M}_{0_j}$ associated with different orientations of the sensor.
Similarly, any spatial environmental changes $\mathbf{e}_{i} \rightarrow \mathbf{e}_{i+1}$ can be considered during the exploration. This means that the new points of view $\mathcal{M}_{i+1_j}$ found by compensation can lie anywhere in the space of points of view and do not necessarily correspond to points of view that the agent already explored.
Although the approach we described to discover metric regularity applies to such generic cases, we propose to carefully select the initial points of view $\mathcal{M}_{0_j}$ and the spatial changes $\mathbf{e}_{i} \rightarrow \mathbf{e}_{i+1}$ in order to limit the computational cost of this second simulation. The goal is to consider the points of view $\mathcal{M}_i \in \big\{\mathcal{M}^*\big\}_0$ estimated in the previous simulation as a given and to ensure that no new point of view needs to be created and stored in memory.
This is done by taking already known points of view as initial points of view,
$$\mathcal{M}_{0_j} = \mathcal{M}_i \in \big\{\mathcal{M}^*\big\}_0 , $$
and by taking all possible displacements in the grid of positions already explored in \ref{sec:Estimating spatially related points of view} as spatial changes $\mathbf{e}_{i} \rightarrow \mathbf{e}_{i+1}$.
This way all initial points of view $\mathcal{M}_{0_j}$, and consequently all the potential $\mathcal{M}_{i+1_j}$found by compensation, belong to the same set $\big\{\mathcal{M}^*\big\}_0$. The simulation thus focuses on a single manifold $\big\{\mathcal{M}^*\big\}$ and how capturing metric regularity changes its structure.
Moreover by considering displacements of the objects inside the grid of positions that originally generated the points of view $\mathcal{M}_{i} \in \big\{\mathcal{M}^*\big\}_0$ (and thus the points of view $\mathcal{M}_{0_j}$), we ensure that any new $\mathcal{M}_{i+1_j}$ found by compensation actually corresponds to an initial point of view  (see Fig.~\ref{fig:regularization}(b)):
\begin{equation}
\text{for all } (i,j) , \text{ there exists } j'\in \{1,\dots,J\} \text{ such that: }
\mathcal{M}_{i+1_j} = \mathcal{M}_{0_j'}.
\end{equation}
This way no new point of view needs to be estimated and stored in memory during the whole simulation. The internal modification required to capture regularity can then be done directly on the already estimated metric $d_\mathcal{M}(\mathcal{M}_i,\mathcal{M}_{i'})$, $\forall (\mathcal{M}_i,\mathcal{M}_{i'}) \in \big\{\mathcal{M}^*\big\}_0$.

In the simulation all possible translations of the object in the grid of positions are experienced by the agent, in no particular order, which ensures that each internal distance $d_\mathcal{M}(\mathcal{M}_i,\mathcal{M}_{i'}), \forall (\mathcal{M}_i,\mathcal{M}_{i'}) \in \big\{\mathcal{M}^*\big\}_0$ is taken into account during the metric regularization process.
Like in previous simulation, we opted for an analytical solution in order to bypass a computationally expensive exploration of the agent.
For any point of view $\mathcal{M}_{i_j}$ and environmental change $\mathbf{e}_{i} \rightarrow \mathbf{e}_{i+1}$, the compensating point of view $\mathcal{M}_{i+1_j}$ is estimated by looking for the point of view in $\big\{\mathcal{M}^*\big\}_0$ which generates the same displacement $[\Delta x^{obj}, \Delta y^{obj}, 0]$ of the sensor (see Eq.~\eqref{eq:displacementobject}).

\subsection{Modifying the internal metric}
\label{sec:Modifying the internal metric}

Following the exploration strategy we just described, for all $i$ the agent is able to discover sets of $J$ points of view changes $\mathcal{M}_{i_j} \rightarrow \mathcal{M}_{i+1_j}$ associated with the same external change $\mathbf{e}_i \rightarrow \mathbf{e}_{i+1}$\footnote{In practice, given the limits of the arm's working space, less than $J$ points of view changes are discovered for each $i$.}.
Yet they correspond to different distances $d_\mathcal{M}(\mathcal{M}_{i_j},\mathcal{M}_{i+1-j})$ in the internal metric estimated by the agent. This phenomenon is illustrated in Fig.~\ref{fig:equalitysets} for three different displacements of the object.
We thus propose to modify this internal metric to comply with the equality constraints inferred from the agent's sensorimotor exploration. 
The target metric $\widehat{d}_\mathcal{M}(\mathcal{M}_{i_j},\mathcal{M}_{i+1_j})$ after regularization should be such that:
\begin{equation}
\widehat{d}_\mathcal{M}(\mathcal{M}_{i_j},\mathcal{M}_{i+1_j}) = D_{i,i+1} \text{ , for all } j
\label{eq:common_distance}
\end{equation}
where $D_{i,i+1}$ is a distance shared by all point of view changes associated with the environmental change $\mathbf{e}_{i} \rightarrow \mathbf{e}_{i+1}$.
Yet the equality constraints discovered by the agent during the exploration do not define the value of $D_{i,i+1}$. We thus propose to derive it from the existing metric defined in the motor space during previous simulation. The distance $D_{i,i+1}$ is set to be equal to the average of the set of distances it is associated with:
\begin{equation}
D_{i,i+1} = \big\langle \: d_\mathcal{M}(\mathcal{M}_{i_j},\mathcal{M}_{i+1_j}) \: \big\rangle_j \:,
\label{eq:distance_equalities}
\end{equation}
where $\langle . \rangle$ denotes the average operator.\\

\textit{Iterative unfolding procedure:}
The regularization method described in Eq.~\eqref{eq:common_distance} and \eqref{eq:distance_equalities} does not necessarily lead to a consistent (low-dimensional) metric. Indeed there is no interplay between the different values $D_{i,i+1}$ associated with different displacements. Consequently two object displacements with a ratio of 2:1 in magnitude are not necessarily associated with internal distances $D_{i,i+1}$ respecting the same ratio.
In order to enforce consistency in the modified metric, we project it in low dimension, the same way the initial metric was projected in Sec.~\ref{sec:Building a topological representation of spatial configuration}.
Forcing the data to lie in a low dimensional space further alters the internal metric by having the different distances $d_\mathcal{M}(\mathcal{M}_{i},\mathcal{M}_{i'})$, for all $(\mathcal{M}_{i},\mathcal{M}_{i'}) \in \big\{\mathcal{M}^*\big\}_0$, interact when positioning each projected data point $\mathcal{M}_{i}$.
The metric obtained after projection is then considered the new internal metric between the points of view estimated by the agent.

The whole process of assessing the values $D_{i,i+1}$, modifying the internal distances $d_\mathcal{M}(\mathcal{M}_{i_j},\mathcal{M}_{i+1_j})$, and projecting the resulting metric in low dimension is repeated $10$ times in order to let it iteratively  converge to a stable equilibrium.
The final metric is expected to both respect the distances equalities inferred during the exploration and be consistent in low dimension.
As in the previous simulation, we independently project the data described by the modified metric in 2D and 3D.\\

\begin{figure*}[t!]
\centering
\includegraphics[width=1\linewidth]{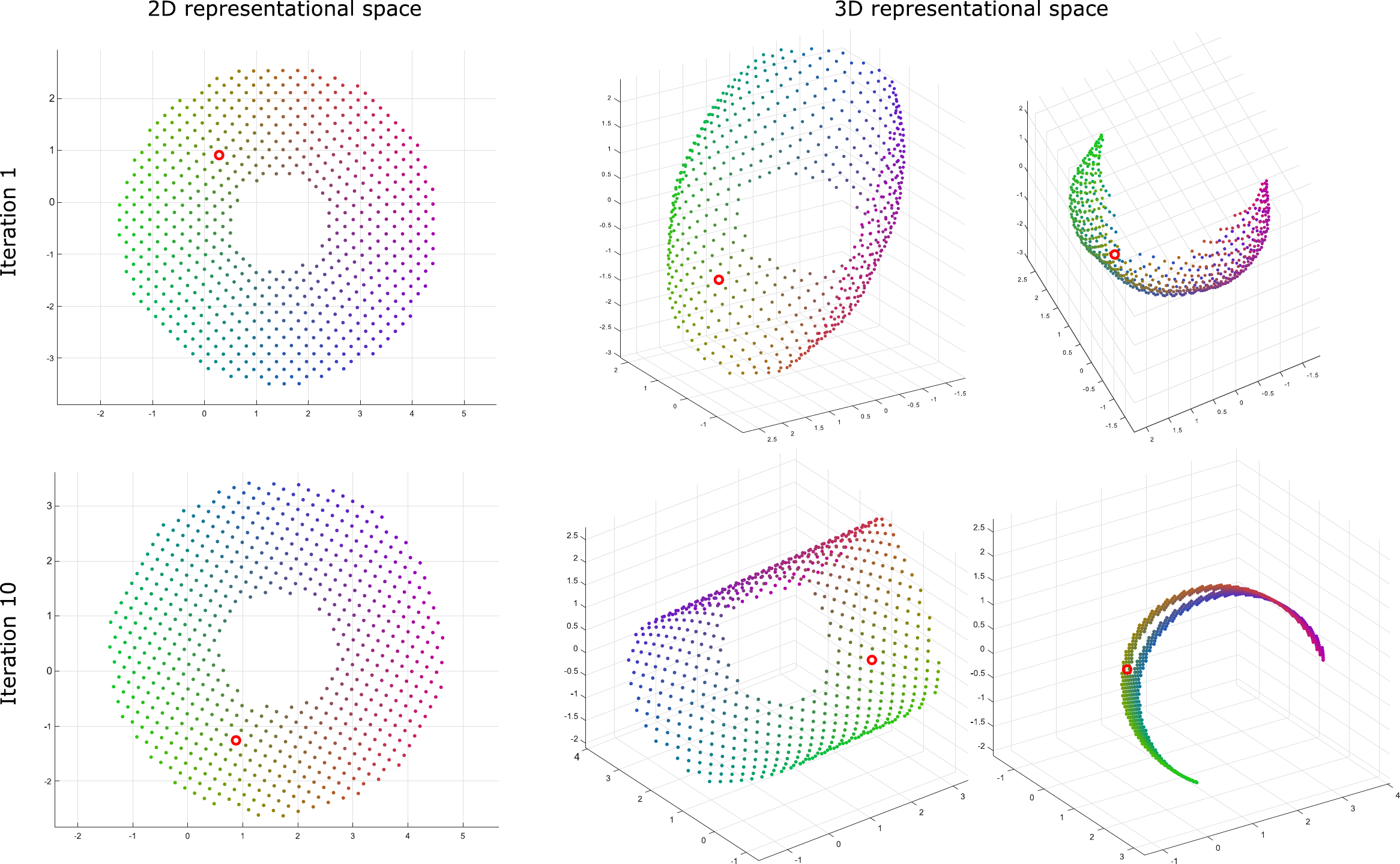}
\caption{Projections of the points of view in 2D and 3D after the first and last iterations of internal metric regularization. A small distortion of the metric in 2D is visible after the first iteration, but disappears after 10 iterations. A large distortion of the metric in 3D is visible after the first iteration.  After 10 iterations, the metric appears regular but the manifold still displays a global curvature.
For the sake of visualization, the missing configuration of Fig.~\ref{fig:compensation_simu1}(a) is displayed in red.}
\label{fig:projections_regularized}
\end{figure*}

\textit{Results:}
Figure~\ref{fig:projections_regularized} presents the 2D and 3D projections of the set $\big\{\mathcal{M}^*\big\}_0$ after the first and last iterations of the regularization process.
Compared to the previous internal representation of Fig.~\ref{fig:projections}, the metric regularization has greatly impacted the shape of the projected manifold. Both 2D and 3D projections still capture the topology of the grid of positions the sensor has visited but now also exhibit a regular metric structure.
Direct neighbors in the internal representations now tend to be equidistant.

\begin{figure}[t!]
\centering
\includegraphics[width=0.5\linewidth]{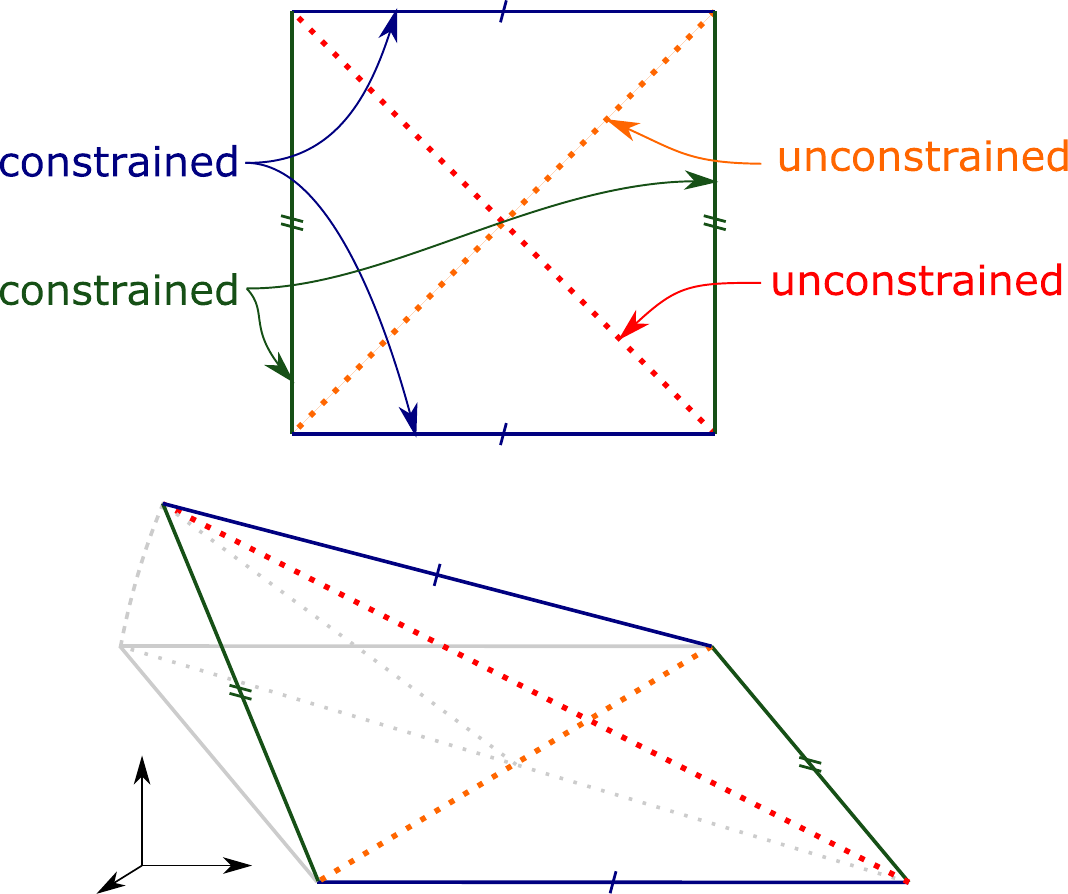} 
\caption{Opposite sides of each square in the grid of points of view are constrained, but not their diagonals. This allows a residual curvature when the representational space is of higher dimension than the manifold's intrinsic dimension.}
\label{fig:diffdiagonals}
\end{figure}

Thanks to the metric regularization, CCA has been able to project the data in 2D without cutting the underlying manifold. Consequently the resulting projection appears to be a good internal representation for the 2D spatial configuration of the agent's sensor. More precisely it is an affine transformation of it, as can be seen in the metric comparison of Fig.~\ref{fig:comparedmetrics}(c). This shows that the internal 2D representation is an excellent representation for its true external counterpart.
This is expected as the manifold of points of view is intrinsically 2-dimensional and thus fits perfectly in a 2D representational space after regularization.
On the other hand, the final 3D projection appears as a slightly folded version of the external grid of positions. Given the additional dimension of the representational space and the equality constraints imposed on the metric, such a seemingly developable manifold is a suitable solution for the projection. Indeed the proposed exploration strategy ensures that the agent discovers that opposite sides of any square in the grid should be equal in length, but does not enforce such a constraint regarding the two diagonals of this square, as illustrated in Fig.~\ref{fig:diffdiagonals}.
As a consequence we can see in Fig.~\ref{fig:comparedmetrics}(c) that the internal 3D projection is a good representation of its external counterpart for small distances but that larger ones are underestimated due to the bending of the manifold.

\subsection{Evaluating the internal representations}
\label{sec:Evaluating the internal representations}

In order to evaluate the benefit of the metric regularization and the quality of the final internal representation of the sensor's spatial configuration, we propose a simple reaching task for the agent to solve.

\begin{figure*}[t!]
\centering
\includegraphics[width=1\linewidth]{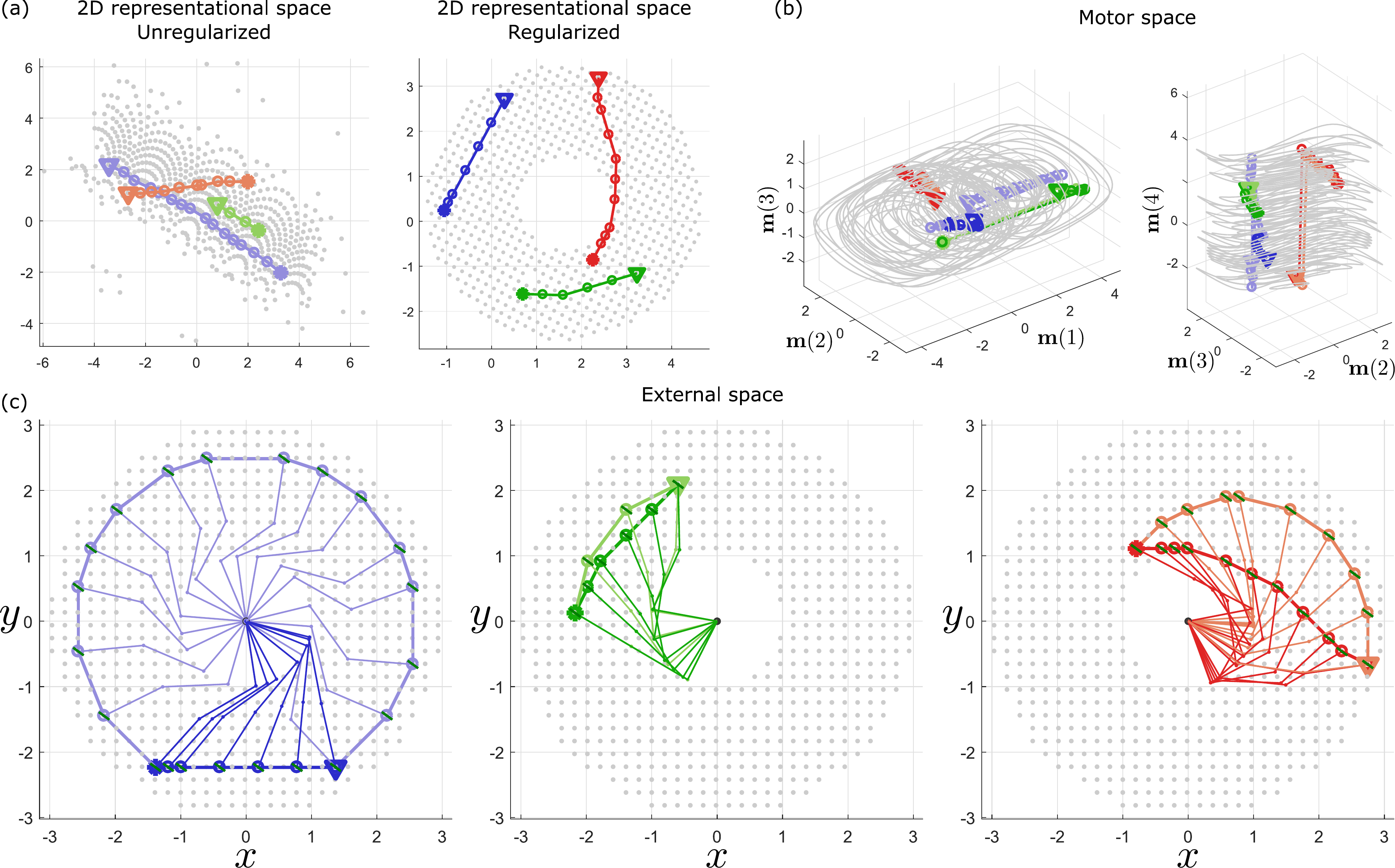}
\caption{Visualization of three reaching trajectories determined by using the 2D representational space, before and after the internal metric regularization. (a) Trajectories in the representational spaces are straight (given the graph constraints they need to respect).
(b) The corresponding motor trajectories sometimes appear to jump around the motor space due to the $2\pi$-periodicity of its metric.
(c) The initial representational space generates curved external trajectories of the sensor {\color{black}(ligh trajectories)}. After regularization of its metric, both internal and external trajectories are straight {\color{black}(darker trajectories)}. (For the sake of visualization, only $10\%$ of the point of view sets are displayed in (b).)}
\label{fig:reaching2D}
\end{figure*}

\begin{figure*}[t!]
\centering
\includegraphics[width=1\linewidth]{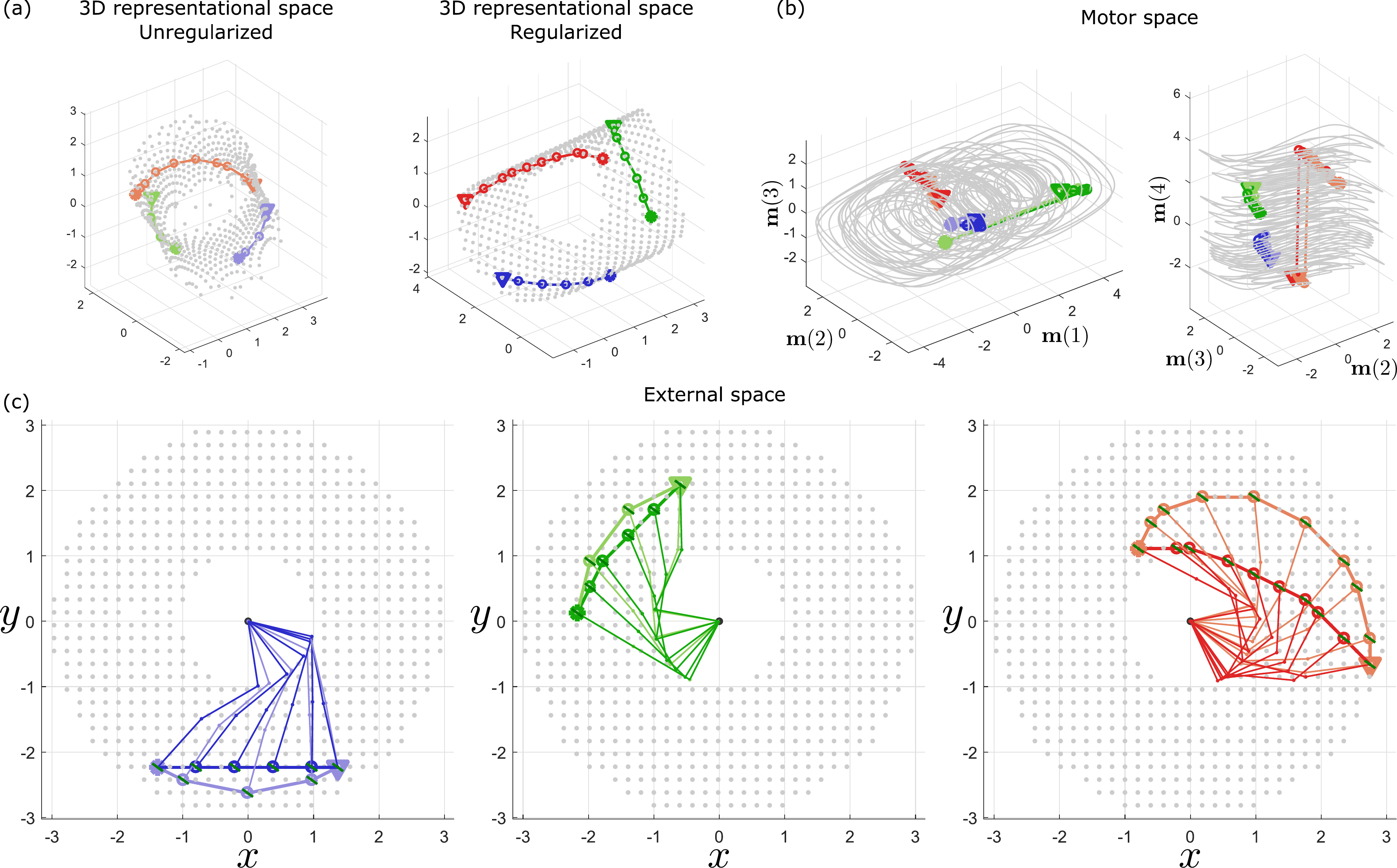}
\caption{Visualization of three reaching trajectories determined by using the 3D representational space, before and after the internal metric regularization. (a) Trajectories in the representational spaces are geodesics (given the graph constraints they need to respect).
(b) The corresponding motor trajectories sometimes appear to jump around the motor space due to the $2\pi$-periodicity of its metric.
(c) The initial representational space generates curved external trajectories of the sensor {\color{black}(light trajectories)}. After regularization of its metric, the trajectories they generate are straight {\color{black}(darker trajectories)}. (For the sake of visualization, only $10\%$ of the point of view sets are displayed in (b).)}
\label{fig:reaching3D}
\end{figure*}

The task consists in finding the shortest path between a starting point of view $\mathcal{M}_s$ and a target point of view $\mathcal{M}_t$, both randomly drawn from $\big\{\mathcal{M}^*\big\}_0$. We are particularly interested in visualizing through which intermediary points of view $\mathcal{M}_p$ the internal representation would require the agent to go during the reaching.
To do so, the set $\big\{\mathcal{M}^*\big\}_0$ is considered a graph in which each $\mathcal{M}_i$ is a node and each $d_\mathcal{M}(\mathcal{M}_i,\mathcal{M}_i')$ corresponds to the length (weight) of the undirected edge between the nodes $i$ and $i'$.
Nevertheless such a fully connected graph contains a direct link between any two nodes, that the agent could take to reach $\mathcal{M}_t$ from $\mathcal{M}_s$ without the need for intermediary steps. We thus prune the fully connected graph to retain only local connections. Any edge of length greater than $d_\mathcal{M}(\mathcal{M}_i,\mathcal{M}_i') = 0.72$ is arbitrarily pruned. This value has been manually set so that each node $\mathcal{M}_i$ is not connected to more than $50$ nodes around it.
This way the search for the shortest path is required to use only local transitions in the internal representation.
Finally the optimal path between $\mathcal{M}_s$ and $\mathcal{M}_t$ is determined by applying Dijkstra's algorithm \citep{Dijkstra1959} on the pruned graph.

In order to keep the visualization uncluttered, only $3$ pairs $\{\mathcal{M}_s,\mathcal{M}_t\}$ are randomly drawn for the evaluation. Four different internal representations are considered to solve the reaching tasks: the 2D and 3D representations before regularization, and the 2D and 3D representations after regularization.

Note that the reaching task as been formalized at the level of points of view. However points of view are an abstraction of the actual motor states the agent controls. In order to know which motor states the agent has to go through during the reaching task, we can look inside each intermediary $\mathcal{M}_p$ for the motor state $\mathbf{m}_p \in \mathcal{M}_p$ which is the closest to previous motor state:
\begin{equation}
\mathbf{m}_p = \argmin_{\mathbf{m}_k \in \mathcal{M}_p} \big( d_\mathbf{m}(\mathbf{m}_k,\mathbf{m}_{p-1}) \big),
\label{eq:reaching_motor}
\end{equation}
with $\mathbf{m}_{0} = \mathbf{m}_{s}$ a starting motor state randomly drawn in $\mathcal{M}_s$.\\

\textit{Results:}
The trajectories found by the agent for the two 2D representations are displayed in Fig.~\ref{fig:reaching2D}. Their 3D counterparts are displayed in Fig.~\ref{fig:reaching3D}.
For each type of representation, we can see that the internal trajectories are topologically consistent with the actual trajectories followed by the sensor in the external space. This indicates that all internal representations correctly captured the topology of the space of spatial configurations of the sensor, as was already discussed in Sec.~\ref{sec:Building a topological representation of spatial configuration}.

\begin{figure*}[t!]
\centering
\includegraphics[width=1\linewidth]{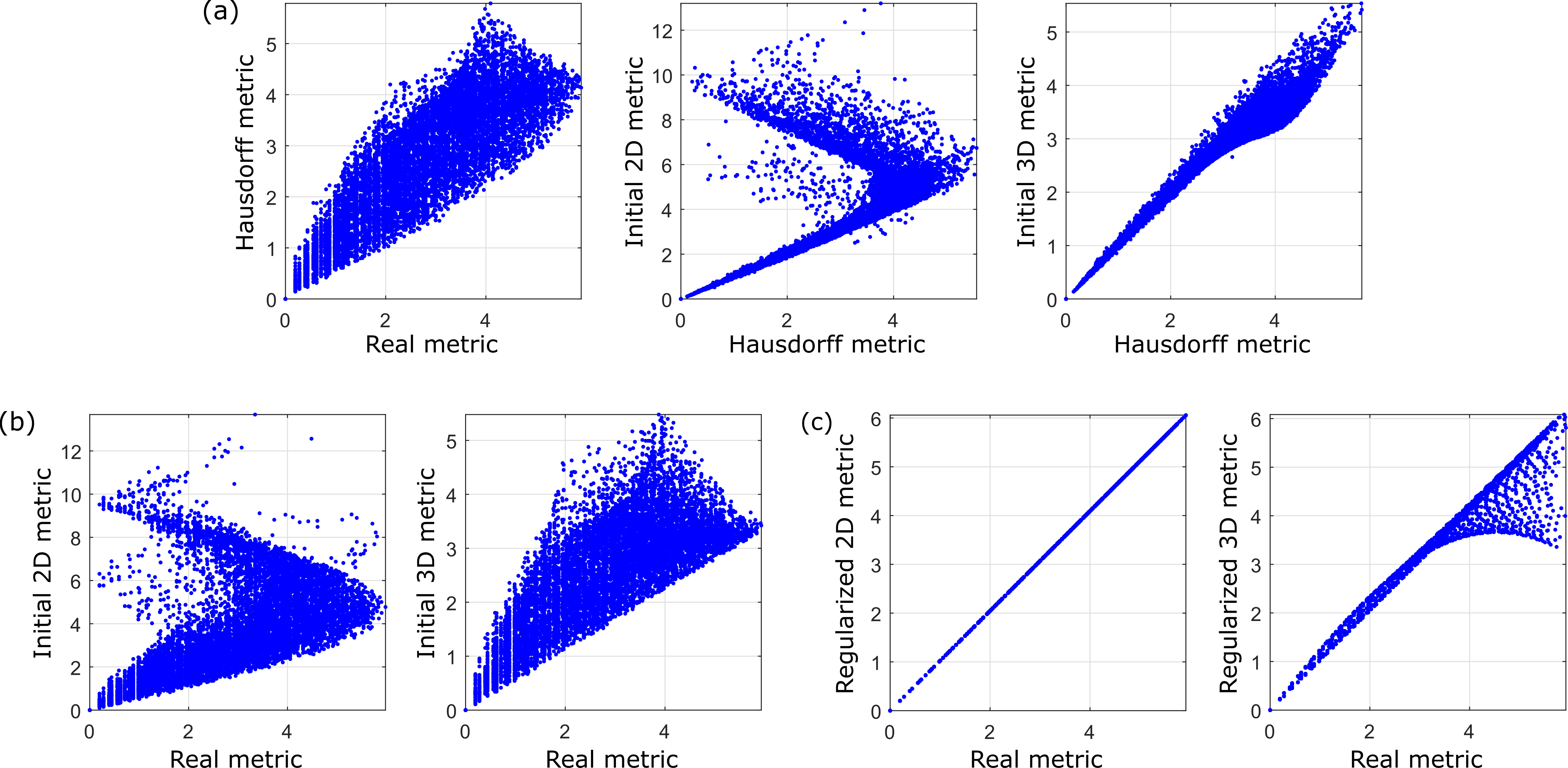}
\caption{Comparisons between the external Euclidean metric computed on the grid of positions taken by the sensor, the Hausdorff metric computed between the points of view, and the internal metrics in the 2D and 3D representational spaces before and after regularization.
{\color{black} Each dot corresponds to a pair of positions taken randomly in the grid of positions (for the sake of visualization, only a randomly selected $5\%$ subset of all possible pairs are displayed, with no qualitative impact on the results). The abscissa and ordinate of each dot in the different plots correspond to the distance between the two positions in the corresponding metrics.}
(a) The Hausdorff metric derived from the motor space greatly differs from the external metric. Moreover the low-dimensional projections in 2D and 3D further distort the original Hausdorff metric. This is particularly true for the 2D projection as we see that CCA was unable to preserve larger distances.
(b) The initial metrics in the 2D and 3D representational spaces differ greatly from the external metric, for both small and large distances. The distortion is even greater for the 2D space due to the impossibility for CCA to correctly unfold the manifold of points of view.
(c) After regularization, the final 2D and 3D metrics are a more satisfying internal representations of the external metric. This is particularly true for the 2D projection which is almost a perfect mapping due to the fact that the manifold is intrinsically 2D. However we can see that the 3D projection tends to underestimate larger distances. This corresponds to the residual curvature of the manifold displayed in Fig.~\ref{fig:projections_regularized}.
}
\label{fig:comparedmetrics}
\end{figure*}

The internal metric regularization however greatly affects the trajectories found by the agent.
By construction the internal trajectories correspond to straight lines (in 2D) and geodesics (in 3D). More precisely, they are as straight as they can be, given the constraint of going through nodes in the grid and the presence of a hole in the center of the working space.
However, before regularization their corresponding external trajectories appear curved. This is due to the difference between the internal metric initially derived from the motor space and the external Euclidean metric.
On the contrary, after regularization, the external trajectories also appear straight. This shows that the regularity of the external Euclidean metric has been correctly captured by the regularized representation.



\section{Discussion}
\label{sec:Discussion}

The primary goal of this work was to define the sensorimotor structure underlying the perception of space by a naive agent.
This includes understanding how spatial knowledge is grounded in the agent's sensorimotor experience, why it is relevant for the agent to extract it, but also how this sensorimotor structure relates to the properties that characterize our subjective perception of space. These properties are that space contains both the agent and its environment, that it is invariant to its content, and that it is isotropic.

Taking inspiration from the SMCT and intuitions by H.Poincar\'e, we proposed to ground spatial knowledge in the way an agent can transform its sensorimotor experience. More precisely this led us to consider \emph{displacements} as a suitable way to characterize space. Fundamentally space is indeed the frame that allows displacement. More interesting, displacements of the agent can also be internally distinguished from other sensorimotor experiences thanks to their sensory redundancy with displacements in the environment (and vice versa). They also support the spatial characteristics we want to capture: both the agent and its environment can undergo displacements, any content of space can be displaced, and displacements can be carried out the same way regardless of the agent's position and orientation.
Our assumption is that the sensory redundancy induced by displacements is worthy for a naive agent to capture. Such knowledge indeed enables it to predict the sensory outcome of some of its motor commands.
It also offers a more compact way to represent the agent's sensorimotor experience. Finally, the structure of the set of displacements an agent can produce is independent of the content of the environment, which ensures some generalization capacities for the agent. Overall, \emph{space} would then be a convenient and useful way for the agent to account for its redundant sensorimotor experiences.

Two simulations have been proposed in this paper to illustrate how a naive agent can discover displacements in its raw sensorimotor experience and how it can derive from it the topology and metric regularity of its spatial configuration.
The simulated system involved a simple robotic arm observing an object made of multiple light sources through an eye-like sensor. By observing displacements of an object and trying to compensate for them from a sensory perspective, the agent was able to isolate the motor changes which are redundant with changes in the environment.
We then proposed to capture the structure underlying these specific motor changes and to build an internal representation of them. This representation was based on the pre-definition of the proto-spatial concept of \textit{point of view} which compacts the agent's experience by grouping together motor states which are equivalent from a sensory perspective (or mechanically redundant from an external perspective).
After the agent experiences the possibility of compensating multiple object displacements from a single initial point of view, the resulting internal representation it builds successfully captures the topology of its sensor's position in the external space.
Moreover, by compensating for multiple object displacements from different initial points of view, we showed how the metric of the internal representation can be modified to capture the metric regularity of space, making the final internal representation a good internal representation for the sensor's external spatial configuration.
Finally we showed how this internal representation can be used to solve simple spatial tasks, like finding the shortest path (both internally and externally) in a reaching task.\\

It is important to notice that the sensorimotor characterization of space we put forward is independent of the particular encoding of the motor and sensory information, and also independent of the structure of the agent's body. Obviously the sensory states experienced by an agent depend on the nature and characteristics of its sensors, and the actions it can perform in the world depend on the nature and characteristics of its motor. Yet the existence of redundant transformations between the agent and its environment is induced by the structure of external space, on which the agent's hardware has no influence. Through its exploration of the world, an agent is thus bound to experience these specific transformations, although they might be encoded differently for different agents. This independence to the encoding and importance of embodiment in the grounding of perceptual experience is directly in line with the SMCT.

Of course this encoding-independence has its limits. One could for example imagine an agent with sensors so simple or exotic that they would destroy or not capture the information needed by the agent to discover space.
More interesting, limited motor capacities could also prevent an agent to fully explore the structure of space. An agent lacking the degrees of freedom to produce motor changes redundant with some displacements in the environment (and thus compensate for them) would not interpret them as displacements but as state changes. The resulting internal representation of space would thus be truncated. In our simulated system, if the object was also able to move along a depth axis (3D space), the agent, still restricted to motions in the plane, would nevertheless build a 2D internal representation of space.
Consequently any displacement of the object orthogonal to this plane would be interpreted as a state change, similarly to the way we perceive rotations of a 4D hypercube as changes in 3D structure.
Although incomplete from an external perspective, this truncated characterization of space is nevertheless adapted and useful for the agent which cannot have any effect in this extra spatial dimension.

In our simulations, we purposefully designed an opposite setting in which we limited the degrees of freedom of objects. Objects were limited to translations in the plane, without the possibility of rotation. Consequently, any rotation of the sensor is not redundant with any potential environmental change.
Although artificial, this setting allowed us to restrain displacements identifiable as such by the agent to an easily visualized set of 2D translations, while rotations of the agents' sensor are interpreted as state changes. The final results demonstrated that rotations have been correctly excluded from the internal representations built by the agent.

Our characterization of space gives the possibility of incomplete representation of space, but it can also lead to the discovery of extra spatial dimensions. This would happen if the agent was able to generate motor changes redundant with some state changes in the environment (and thus to compensate for them). One can for instance imagine an agent able to control the sensitivity of its sensors, which would then be redundant with an overall change of light intensity in the environment. This redundant state change would thus be incorporated in the internal representation of the agent as an extra "spatial" dimension. Although incorrect from an external perspective, this representation would yet again be adapted and useful for the agent which can compensate for these specific state changes as if they were displacements.

Note also that even in a standard setting -- like our simulations in which the agent correctly discovers the expected spatial dimensions -- the internal representation of space cannot capture all properties of space. Indeed some of them like its unit of length or its orientation are arbitrarily defined by an external observer. Consequently the scale and orientation of the internal representation is different from its external counterpart (grid of positions). The agent has no means to access these properties, but also no incentive to do so as they have no relevance to its ability to change its spatial configuration.\\

The general approach and simulations we proposed in this paper give some insight into the sensorimotor grounding of the concept of space. Yet they also have multiple limitations.
The agent-environment system we considered is relatively simple.
The agent is a single serial arm equipped with a single low dimensional sensor. Although the concept of redundancy/compensability holds for more complex agents (see for instance an agent equipped with two independent moving eyes and two ears in \citep{Laflaquiere2010}), the occurrence of compensation experiences appears less probable, and the tracking-like behavior more difficult to achieve.
For instance, imagine the agent was made of two arms instead of one. Compensable experiences would consist in jointly moving both arm tips in a rigid fashion to follow the displacements of an object. On the contrary, moving a single arm would be interpreted as a state change.
Thus compensable experiences are not impossible but more difficult to encounter when the agent gets more complex. An interesting solution to this problem could be for the agent to build multiple local notions of space (one for each arm in the previous example) by experiencing redundancy/compensability in subspaces of its whole sensory space. At a later developmental stage, all these local descriptions of space could be united in a single consistent internal representation which would capture information about both space and the structure of the agent's body.

The simplicity of the agent mechanical structure also allowed us to implement a method for estimating basic points of view. It takes advantage of the fact that the arm has a single mechanical degree of redundancy and hinge joints, making each point of view a closed 1D manifold. The method would however not directly scale up to manifolds of greater dimensions and/or with boundaries.

The environment considered in the simulations is also very simple; it comes down to a single object made up of rigidly connected light sources. This unrealistic setup was proposed for the simplicity of the simulation, and to guarantee the possibility of compensation in the system. In the presence of a more complex environment made up of multiple objects, the displacement of a single object would generate a sensory variation that the agent could not entirely compensate. Indeed the tracking of the moving object would also lead to a relative motion of all the other objects that did not move.
A potential solution to adapt the approach to more realistic environments would be to identify redundancy/compensability in sensory subspaces that capture information about only a subpart of the environment (similar to a receptive field in the retina). This way the agent could track the moving object in such a subspace, while discarding the variations it generates in the rest of the sensory space during its exploration.
Another solution would consist in assuming that the environment is static most of the time but that the agent can move its base in space, the same way we can move on our feet. Such a displacement of the base would be equivalent, from a sensory perspective, to an opposite displacement of the whole environment that the agent could try to compensate by moving the rest of its body. This would however suggest that only agents capable of moving their base in the world are able to develop a notion of space. To avoid this unnecessary limitation, a mix of both solutions appears more promising.

The methods we proposed in this paper are partly based on analytic solutions which rely on the knowledge of the arm's forward model and Jacobian, as well as the actual displacement of the object in space.
The primary purpose of the computational method implemented in these simulations is thus illustrative. They are unrealistic from a developmental standpoint in which these information are not accessible to the naive agent. One could however consider possible solutions to avoid these analytic shortcuts.
%
They would require the development of exploration strategies that efficiently discover motor states which always generate equivalent sensory states, or those which compensate for a given sensory change. The building of the internal representation could also be made more realistic by using for example a neural network like was already proposed in \citep{Laflaquiere2013}.
Such approaches should also be made robust to more realistic exploration scenarios in which the agent cannot necessarily sample its whole working space between two displacements of the object. This should not present a challenge, as our approach is theoretically able to accommodate itself to an incremental discovery of all displacements. The only fundamental assumption that needs to be respected is that the agent can statistically move more often than the environmental state changes. Otherwise it would be impossible for the agent to estimate the effect of its actions on a sensory flow which is constantly changing at a rapid rate due to the environment.

It is also important to notice that, although internal representations were built in the two simulations of this paper, we do not claim that such explicit representations are necessary for an agent to interact with its environment. Indeed all relevant information about space topology and metric regularity are contained in the internal metric derived from the motor space and modified by the redundancies discovered by the agent. The metric knowledge is sufficient to define the manifolds the agent has to capture without the need for low-dimensional embedding. This is for instance illustrated in section~\ref{sec:Evaluating the internal representations} in which only the knowledge of the metric $d_\mathcal{M}(\mathcal{M},\mathcal{M})$, representing the estimated distances between underlying sets of motor states, is used to solve the reaching task.
{\color{black} Thus, beyond their illustrative power, we do not argue that the low-dimensional representations built in this work are necessary for spatial knowledge. We rather think that the predictive capacity supported by the sensorimotor invariants is constitutive of this spatial knowledge (for instance, knowing that a set of motor commands would generate the same sensory input as the one currently experienced).}
In our second simulation, the explicit embedding of the metric in low dimension has however been useful to enforce consistency (see section \ref{sec:Modifying the internal metric}). Other ways to regularize the metric could nonetheless be considered. For instance by observing rotations as well as successive displacements of objects, it might be possible for the agent to discover relations between the different distances that constitute its internal metric, without the need for a low-dimensional embedding that the agent has a priori no incentive to produce. Interestingly such a regularization would also solve the problem of inconsistent diagonal lengths that was noticed in Fig.~\ref{fig:diffdiagonals} and led to a curved internal representation in Fig.\ref{fig:projections_regularized}.

Although desirable to fully regularize the internal metric, the addition of rotations in the system introduces other problems.
In this paper rotations have artificially been kept out of the simulations by allowing only translations of the object. Consequently any rotation of the sensor is considered a state change, due to a lack of compensability, and ignored in the building of the internal representation. Yet in a more realistic setting, rotations in the plane are also compensable and should therefore also be considered displacements in a mathematical sense. This has been partly illustrated in \citep{Laflaquiere2015a} in which a similar agent captures both translations and rotations of its sensor in a 3D internal representation.
The low-dimensional projection of such a 3D manifold is however made difficult by the looped third dimension induced by rotations. It makes it impossible to project and visualize the data in 3D, although it corresponds to the manifold's intrinsic dimension.
But more importantly, the addition of rotations would disrupt the metric regularization method we proposed in this paper. Indeed for a rotation of the object, the amplitude of the compensating displacement of the sensor depends on the distance to the center of rotation. Thus the regularization of the metric along the dimension related to rotations would not be straightforward.
A solution to this problem could be to distinguish rotations from translations, based on their different properties. Rotations and translations for instance do not follow the same compositionality rules: the order of translations in a sequence of transformations does not affect the end result, which is not true for rotations. Another property is that dimensions associated with rotations loop on themselves. Finally, unlike translations, rotations keep points unchanged in space (the axis of rotation). These different properties relate to our subjective experience of rotations. Studying how they manifest themselves in an agent's sensorimotor experience might be the way to better capture them in the internal representation of spatial configuration.\\

Beyond the several potential improvements of the current simulations we just mentioned, future work on the sensorimotor grounding of space perception has to answer many important questions. In our approach spatial knowledge is fundamentally rooted in the motor space. As presented in this paper, it provides a characterization of the \emph{agent}'s spatial configuration. Yet our spatial perception extends past our own body to include our environment. Thus in order to reach a more complete perception of space, it is necessary to develop the sensorimotor framework to account for the spatial configuration of objects in the environment. As proposed by H.Poincar\'e \citep{Poincare1895}, this knowledge should fundamentally be interpreted by the agent via its own capacity to generate spatial changes. The distance to an object would for instance be internally encoded as the motor command needed to reach it. This way the agent can ground these external properties it cannot directly access in its motor space, which it can access and control.
In order to fully answer this question, one will probably have to also tackle the problem of object perception and how it can co-emerge with the notion of space. Some preliminary steps have already been done in this direction \citep{laflaquiere2015objects,Laflaquiere2016}.
{\color{black} Only once the grounding of spatial knowledge presented in this work has been extended to the spatial configuration of the environment will it be possible to consider complex applications for robotics, like navigation or object manipulation.}

Finally our long term objective is to let a robot build its own grounded knowledge of how its body and objects around it behave in space{\color{black}, first by theoretically defining the type of sensorimotor structure which supports this knowledge and second by developing algorithmic solutions to capture this structure}.
This {\color{black}spatial} common-sense knowledge could then be re-used in an open-ended manner to efficiently solve unforeseen spatial tasks the robot would face in the world.
Nonetheless, beyond this technical motivation, we also aim to shed some light on the fundamental nature of space and how our perception of space emerges. We try to trace back the origin of our subjective experience of space and explicitly define how its peculiar properties manifest themselves in our sensorimotor experience.

{\color{black} On a more philosophical side, the approach developed in this paper suggests that the concept of space might emerge from sensorimotor interactions with the environment, without the need for specific spatial prior knowledge. In that sense, and within the current discussion about the importance of priors in artificial intelligence \cite{lake2017building, marcus2018deep, sunderhauf2018limits}, our approach suggests the possibility of a fully data-driven emergence of perception. However this claim needs to be put in perspective, since the mechanisms that would lead an agent to capture the sensorimotor invariants supporting spatial knowledge are not yet fully understood (in particular from an algorithmic point of view). Nevertheless we aim at pushing the SMCT-based approach we have developed as far as possible using the assumption of minimal prior knowledge. This has also led us to challenge the idea that the convolution mechanism used in today's successful convolutional neural networks, and often considered an argument for the necessity of priors, has to be pre-implemented in the perceptive system \cite{Laflaquiere2016}. Regardless of the unfolding of this theoretical debate and its consequence for the field of artificial perception, it is nonetheless probable that natural selection has endowed us with particular priors to bootstrap and speed up the development of our perceptive abilities. This would of course include priors about the concept of space, which would certainly be very useful due to the pervasiveness of space in our perception of the world. On the other hand, the presence of such priors in our brain, proven or not, does not solve the problem of roboticists who still need to identify and implement them in their robots if they want to accelerate their development.}


\begin{thebibliography}{51}
\providecommand{\natexlab}[1]{#1}
\providecommand{\url}[1]{\texttt{#1}}
\expandafter\ifx\csname urlstyle\endcsname\relax
  \providecommand{\doi}[1]{doi: #1}\else
  \providecommand{\doi}{doi: \begingroup \urlstyle{rm}\Url}\fi

\bibitem[Bengio et~al.(2013)Bengio, Courville, and Vincent]{Bengio2012}
Yoshua Bengio, Aaron Courville, and Pascal Vincent.
\newblock Representation learning: A review and new perspectives.
\newblock \emph{IEEE transactions on pattern analysis and machine
  intelligence}, 35\penalty0 (8):\penalty0 1798--1828, 2013.

\bibitem[Cadena et~al.(2016)Cadena, Carlone, Carrillo, Latif, Scaramuzza,
  Neira, Reid, and Leonard]{Cadena2016}
Cesar Cadena, Luca Carlone, Henry Carrillo, Yasir Latif, Davide Scaramuzza,
  Jos{\'e} Neira, Ian Reid, and John~J Leonard.
\newblock Past, present, and future of simultaneous localization and mapping:
  Toward the robust-perception age.
\newblock \emph{IEEE Transactions on Robotics}, 32\penalty0 (6):\penalty0
  1309--1332, 2016.

\bibitem[Choe and Smith(2006)]{choe2006motion}
Yoonsuck Choe and Noah~H Smith.
\newblock Motion-based autonomous grounding: Inferring external world
  properties from encoded internal sensory states alone.
\newblock In \emph{Proceedings of the National Conference on Artificial
  Intelligence}, volume~21, page 936. Menlo Park, CA; Cambridge, MA; London;
  AAAI Press; MIT Press; 1999, 2006.

\bibitem[Dearden et~al.(2005)Dearden, Demiris, Kaelbling, and
  Saffotti]{Dearden2005}
Anthony Dearden, Yiannis Demiris, LP~Kaelbling, and A~Saffotti.
\newblock Learning forward models for robots.
\newblock In \emph{IJCAI-INT JOINT CONF ARTIF INTELL}, pages 1440--1445, 2005.

\bibitem[Demartines and H{\'e}rault(1997)]{Demartines1997}
Pierre Demartines and Jeanny H{\'e}rault.
\newblock Curvilinear component analysis: A self-organizing neural network for
  nonlinear mapping of data sets.
\newblock \emph{IEEE Transactions on neural networks}, 8\penalty0 (1):\penalty0
  148--154, 1997.

\bibitem[Dijkstra(1959)]{Dijkstra1959}
Edsger~W Dijkstra.
\newblock A note on two problems in connexion with graphs.
\newblock \emph{Numerische mathematik}, 1\penalty0 (1):\penalty0 269--271,
  1959.

\bibitem[Friston(2010)]{Friston2010}
Karl Friston.
\newblock The free-energy principle: a unified brain theory?
\newblock \emph{Nature Reviews Neuroscience}, 11\penalty0 (2):\penalty0
  127--138, 2010.

\bibitem[Goodfellow et~al.(2016)Goodfellow, Bengio, and
  Courville]{Goodfellow-et-al-2016}
Ian Goodfellow, Yoshua Bengio, and Aaron Courville.
\newblock \emph{Deep Learning}.
\newblock MIT Press, 2016.
\newblock \url{http://www.deeplearningbook.org}.

\bibitem[Greydanus et~al.(2017)Greydanus, Koul, Dodge, and
  Fern]{greydanus2017visualizing}
Sam Greydanus, Anurag Koul, Jonathan Dodge, and Alan Fern.
\newblock Visualizing and understanding atari agents.
\newblock \emph{arXiv preprint arXiv:1711.00138}, 2017.

\bibitem[Hemion(2016)]{Hemion2017}
Nikolas~J Hemion.
\newblock Context discovery for model learning in partially observable
  environments.
\newblock In \emph{Development and Learning and Epigenetic Robotics
  (ICDL-EpiRob), 2016 Joint IEEE International Conference on}, pages 294--299.
  IEEE, 2016.

\bibitem[Hoffman et~al.(2015)Hoffman, Singh, and Prakash]{Hoffman2015}
Donald~D Hoffman, Manish Singh, and Chetan Prakash.
\newblock The interface theory of perception.
\newblock \emph{Psychonomic bulletin \& review}, 22\penalty0 (6):\penalty0
  1480--1506, 2015.

\bibitem[Hohwy(2016)]{Hohwy2016}
Jakob Hohwy.
\newblock The self-evidencing brain.
\newblock \emph{No{\^u}s}, 50\penalty0 (2):\penalty0 259--285, 2016.

\bibitem[Jonschkowski and Brock(2015)]{Jonschkowski2015}
Rico Jonschkowski and Oliver Brock.
\newblock Learning state representations with robotic priors.
\newblock \emph{Autonomous Robots}, 39\penalty0 (3):\penalty0 407--428, 2015.

\bibitem[Kant(1998)]{kant1998critique}
Immanuel Kant.
\newblock \emph{Critique of pure reason}.
\newblock Cambridge University Press, 1998.

\bibitem[Laflaquiere(2017)]{Laflaquiere2016}
Alban Laflaquiere.
\newblock Grounding the experience of a visual field through sensorimotor
  contingencies.
\newblock \emph{Neurocomputing}, 2017.

\bibitem[Laflaquiere and Hemion(2015)]{laflaquiere2015objects}
Alban Laflaquiere and Nikolas Hemion.
\newblock Grounding object perception in a naive agent's sensorimotor
  experience.
\newblock In \emph{Development and Learning and Epigenetic Robotics
  (ICDL-EpiRob), 2015 Joint IEEE International Conference on}, pages 276--282.
  IEEE, 2015.

\bibitem[Laflaquiere et~al.(2010)Laflaquiere, Argentieri, Gas, and
  Castillo-Castenada]{Laflaquiere2010}
Alban Laflaquiere, Sylvain Argentieri, Bruno Gas, and Eduardo
  Castillo-Castenada.
\newblock Space dimension perception from the multimodal sensorimotor flow of a
  naive robotic agent.
\newblock In \emph{Intelligent Robots and Systems (IROS), 2010 IEEE/RSJ
  International Conference on}, pages 1520--1525. IEEE, 2010.

\bibitem[Laflaquiere et~al.(2012)Laflaquiere, Argentieri, Breysse, Genet, and
  Gas]{Laflaquiere2012}
Alban Laflaquiere, Sylvain Argentieri, Olivia Breysse, St{\'e}phane Genet, and
  Bruno Gas.
\newblock A non-linear approach to space dimension perception by a naive agent.
\newblock In \emph{Intelligent Robots and Systems (IROS), 2012 IEEE/RSJ
  International Conference on}, pages 3253--3259. IEEE, 2012.

\bibitem[Laflaquiere et~al.(2013)Laflaquiere, Terekhov, Gas, and
  O'Regan]{Laflaquiere2013}
Alban Laflaquiere, Alexander~V Terekhov, Bruno Gas, and J~Kevin O'Regan.
\newblock Learning an internal representation of the end-effector configuration
  space.
\newblock In \emph{Intelligent Robots and Systems (IROS), 2013 IEEE/RSJ
  International Conference on}, pages 1230--1235. IEEE, 2013.

\bibitem[Laflaquiere et~al.(2015)Laflaquiere, Hemion, Ortiz, and
  Baillie]{Laflaquiere2015}
Alban Laflaquiere, Nikolas Hemion, Michael~Garcia Ortiz, and Jean-Christophe
  Baillie.
\newblock Grounding perception: A developmental approach to sensorimotor
  contingencies.
\newblock In \emph{IEEE/RSJ Int. Conf. on Intelligent Robots and Systems},
  pages 1--8, 2015.

\bibitem[Laflaqui{\`e}re et~al.(2015)Laflaqui{\`e}re, O’Regan, Argentieri,
  Gas, and Terekhov]{Laflaquiere2015a}
Alban Laflaqui{\`e}re, J~Kevin O’Regan, Sylvain Argentieri, Bruno Gas, and
  Alexander~V Terekhov.
\newblock Learning agent’s spatial configuration from sensorimotor
  invariants.
\newblock \emph{Robotics and Autonomous Systems}, 71:\penalty0 49--59, 2015.

\bibitem[Lake et~al.(2017)Lake, Ullman, Tenenbaum, and
  Gershman]{lake2017building}
Brenden~M Lake, Tomer~D Ullman, Joshua~B Tenenbaum, and Samuel~J Gershman.
\newblock Building machines that learn and think like people.
\newblock \emph{Behavioral and Brain Sciences}, 40, 2017.

\bibitem[LeCun et~al.(2015)LeCun, Bengio, and Hinton]{LeCun2015}
Yann LeCun, Yoshua Bengio, and Geoffrey Hinton.
\newblock Deep learning.
\newblock \emph{Nature}, 521\penalty0 (7553):\penalty0 436--444, 2015.

\bibitem[Lee and Verleysen(2007)]{Lee2007}
John~A Lee and Michel Verleysen.
\newblock \emph{Nonlinear dimensionality reduction}.
\newblock Springer Science \& Business Media, 2007.

\bibitem[Levine et~al.(2016)Levine, Finn, Darrell, and Abbeel]{levine2016end}
Sergey Levine, Chelsea Finn, Trevor Darrell, and Pieter Abbeel.
\newblock End-to-end training of deep visuomotor policies.
\newblock \emph{Journal of Machine Learning Research}, 17\penalty0
  (39):\penalty0 1--40, 2016.

\bibitem[Lillicrap et~al.(2015)Lillicrap, Hunt, Pritzel, Heess, Erez, Tassa,
  Silver, and Wierstra]{Lillicrap2015}
Timothy~P Lillicrap, Jonathan~J Hunt, Alexander Pritzel, Nicolas Heess, Tom
  Erez, Yuval Tassa, David Silver, and Daan Wierstra.
\newblock Continuous control with deep reinforcement learning.
\newblock \emph{arXiv preprint arXiv:1509.02971}, 2015.

\bibitem[Loviken and Hemion(2017)]{Loviken2017}
Pontus Loviken and Nikolas Hemion.
\newblock Online-learning and planning in high dimensions with finite element
  goal babbling.
\newblock \emph{7th Joint International Conference on Development and Learning
  and Epigenetic Robotics, ICDL-EpiRob 2017}, 2017.

\bibitem[Marcus(2018)]{marcus2018deep}
Gary Marcus.
\newblock Deep learning: A critical appraisal.
\newblock \emph{arXiv preprint arXiv:1801.00631}, 2018.

\bibitem[Mirowski et~al.(2016)Mirowski, Pascanu, Viola, Soyer, Ballard, Banino,
  Denil, Goroshin, Sifre, Kavukcuoglu, et~al.]{Mirowski2016}
Piotr Mirowski, Razvan Pascanu, Fabio Viola, Hubert Soyer, Andy Ballard, Andrea
  Banino, Misha Denil, Ross Goroshin, Laurent Sifre, Koray Kavukcuoglu, et~al.
\newblock Learning to navigate in complex environments.
\newblock \emph{arXiv preprint arXiv:1611.03673}, 2016.

\bibitem[Mnih et~al.(2015)Mnih, Kavukcuoglu, Silver, Rusu, Veness, Bellemare,
  Graves, Riedmiller, Fidjeland, Ostrovski, et~al.]{Mnih2015}
Volodymyr Mnih, Koray Kavukcuoglu, David Silver, Andrei~A Rusu, Joel Veness,
  Marc~G Bellemare, Alex Graves, Martin Riedmiller, Andreas~K Fidjeland, Georg
  Ostrovski, et~al.
\newblock Human-level control through deep reinforcement learning.
\newblock \emph{Nature}, 518\penalty0 (7540):\penalty0 529--533, 2015.

\bibitem[Modayil and Kuipers(2008)]{modayil2008initial}
Joseph Modayil and Benjamin Kuipers.
\newblock The initial development of object knowledge by a learning robot.
\newblock \emph{Robotics and autonomous systems}, 56\penalty0 (11):\penalty0
  879--890, 2008.

\bibitem[Nicod(1924)]{nicod1965geometrie}
Jean Nicod.
\newblock \emph{La g{\'e}om{\'e}trie dans le monde sensible}.
\newblock Presses universitaires de France, 1924.

\bibitem[O'Regan(2011)]{ORegan2012}
J~Kevin O'Regan.
\newblock \emph{Why red doesn't sound like a bell: Understanding the feel of
  consciousness}.
\newblock Oxford University Press, 2011.

\bibitem[O'Regan and No{\"e}(2001)]{ORegan2001}
J~Kevin O'Regan and Alva No{\"e}.
\newblock A sensorimotor account of vision and visual consciousness.
\newblock \emph{Behavioral and brain sciences}, 24\penalty0 (5):\penalty0
  939--973, 2001.

\bibitem[Pan and Yang(2010)]{pan2010survey}
Sinno~Jialin Pan and Qiang Yang.
\newblock A survey on transfer learning.
\newblock \emph{IEEE Transactions on knowledge and data engineering},
  22\penalty0 (10):\penalty0 1345--1359, 2010.

\bibitem[Philipona et~al.(2003)Philipona, O'Regan, and Nadal]{Philipona2003}
David Philipona, J~Kevin O'Regan, and J-P Nadal.
\newblock Is there something out there? inferring space from sensorimotor
  dependencies.
\newblock \emph{Neural computation}, 15\penalty0 (9):\penalty0 2029--2049,
  2003.

\bibitem[Philipona et~al.(2004)Philipona, O'regan, Nadal, and
  Coenen]{Philipona2004}
David Philipona, Jk~O'regan, J-P Nadal, and Olivier Coenen.
\newblock Perception of the structure of the physical world using unknown
  multimodal sensors and effectors.
\newblock In \emph{Advances in neural information processing systems}, pages
  945--952, 2004.

\bibitem[Philipona and O'regan(2006)]{Philipona2006}
David~L Philipona and J~Kevin O'regan.
\newblock Color naming, unique hues, and hue cancellation predicted from
  singularities in reflection properties.
\newblock \emph{Visual neuroscience}, 23\penalty0 (3-4):\penalty0 331--339,
  2006.

\bibitem[Pierce and Kuipers(1997)]{pierce1997map}
David Pierce and Benjamin~J Kuipers.
\newblock Map learning with uninterpreted sensors and effectors.
\newblock \emph{Artificial Intelligence}, 92\penalty0 (1-2):\penalty0 169--227,
  1997.

\bibitem[Poincar{\'e}(1895)]{Poincare1895}
Henri Poincar{\'e}.
\newblock L'espace et la g{\'e}om{\'e}trie.
\newblock \emph{Revue de m{\'e}taphysique et de morale}, 3\penalty0
  (6):\penalty0 631--646, 1895.

\bibitem[Rolf et~al.(2010)Rolf, Steil, and Gienger]{Rolf2010}
Matthias Rolf, Jochen~J Steil, and Michael Gienger.
\newblock Goal babbling permits direct learning of inverse kinematics.
\newblock \emph{IEEE Transactions on Autonomous Mental Development}, 2\penalty0
  (3):\penalty0 216--229, 2010.

\bibitem[Rusu et~al.(2016)Rusu, Rabinowitz, Desjardins, Soyer, Kirkpatrick,
  Kavukcuoglu, Pascanu, and Hadsell]{rusu2016progressive}
Andrei~A Rusu, Neil~C Rabinowitz, Guillaume Desjardins, Hubert Soyer, James
  Kirkpatrick, Koray Kavukcuoglu, Razvan Pascanu, and Raia Hadsell.
\newblock Progressive neural networks.
\newblock \emph{arXiv preprint arXiv:1606.04671}, 2016.

\bibitem[Seth(2014)]{Seth2014}
Anil~K Seth.
\newblock A predictive processing theory of sensorimotor contingencies:
  Explaining the puzzle of perceptual presence and its absence in synesthesia.
\newblock \emph{Cognitive neuroscience}, 5\penalty0 (2):\penalty0 97--118,
  2014.

\bibitem[Shwartz-Ziv and Tishby(2017)]{Shwartz-Ziv1999}
Ravid Shwartz-Ziv and Naftali Tishby.
\newblock Opening the black box of deep neural networks via information.
\newblock \emph{arXiv preprint arXiv:1703.00810}, 2017.

\bibitem[Siciliano and Khatib(2016)]{siciliano2016springer}
Bruno Siciliano and Oussama Khatib.
\newblock \emph{{Springer handbook of robotics}}.
\newblock Springer, 2016.

\bibitem[S{\"u}nderhauf et~al.(2018)S{\"u}nderhauf, Brock, Scheirer, Hadsell,
  Fox, Leitner, Upcroft, Abbeel, Burgard, Milford,
  et~al.]{sunderhauf2018limits}
Niko S{\"u}nderhauf, Oliver Brock, Walter Scheirer, Raia Hadsell, Dieter Fox,
  J{\"u}rgen Leitner, Ben Upcroft, Pieter Abbeel, Wolfram Burgard, Michael
  Milford, et~al.
\newblock The limits and potentials of deep learning for robotics.
\newblock \emph{The International Journal of Robotics Research}, 37\penalty0
  (4-5):\penalty0 405--420, 2018.

\bibitem[Tenenbaum et~al.(2000)Tenenbaum, De~Silva, and
  Langford]{Tenenbaum2000}
Joshua~B Tenenbaum, Vin De~Silva, and John~C Langford.
\newblock A global geometric framework for nonlinear dimensionality reduction.
\newblock \emph{science}, 290\penalty0 (5500):\penalty0 2319--2323, 2000.

\bibitem[Terekhov and O'Regan(2013)]{Terekhov2013}
Alexander~V Terekhov and J~Kevin O'Regan.
\newblock Space as an invention of biological organisms.
\newblock \emph{arXiv preprint arXiv:1308.2124}, 2013.

\bibitem[Terekhov and O'Regan(2014)]{Terekhov2014}
Alexander~V Terekhov and J~Kevin O'Regan.
\newblock Learning abstract perceptual notions: The example of space.
\newblock In \emph{Development and Learning and Epigenetic Robotics
  (ICDL-Epirob), 2014 Joint IEEE International Conferences on}, pages 368--373.
  IEEE, 2014.

\bibitem[Witzel et~al.(2015)Witzel, Cinotti, and O'regan]{Witzel2015}
Christoph Witzel, Fran{\c{c}}ois Cinotti, and J~Kevin O'regan.
\newblock What determines the relationship between color naming, unique hues,
  and sensory singularities: Illuminations, surfaces, or photoreceptors?
\newblock \emph{Journal of vision}, 15\penalty0 (8):\penalty0 19--19, 2015.

\bibitem[Zahavy et~al.(2016)Zahavy, Ben-Zrihem, and Mannor]{Zahavy2016}
Tom Zahavy, Nir Ben-Zrihem, and Shie Mannor.
\newblock Graying the black box: Understanding dqns.
\newblock In \emph{International Conference on Machine Learning}, pages
  1899--1908, 2016.

\end{thebibliography}
\end{document}